\PassOptionsToPackage{unicode}{hyperref}
\PassOptionsToPackage{hyphens}{url}
\documentclass[
]{article}
\usepackage{geometry}
\usepackage{amsmath}
\usepackage{amssymb}
\usepackage{lmodern}
\usepackage{iftex}
\ifPDFTeX
  \usepackage[T1]{fontenc}
  \usepackage[utf8]{inputenc}
  \usepackage{textcomp} 
\else 
  \usepackage{unicode-math}
  \defaultfontfeatures{Scale=MatchLowercase}
  \defaultfontfeatures[\rmfamily]{Ligatures=TeX,Scale=1}
\fi
\IfFileExists{upquote.sty}{\usepackage{upquote}}{}
\IfFileExists{microtype.sty}{
  \usepackage[]{microtype}
  \UseMicrotypeSet[protrusion]{basicmath} 
}{}
\makeatletter
\@ifundefined{KOMAClassName}{
  \IfFileExists{parskip.sty}{%
    \usepackage{parskip}
  }{
    \setlength{\parindent}{0pt}
    \setlength{\parskip}{6pt plus 2pt minus 1pt}}
}{
  \KOMAoptions{parskip=half}}
\makeatother
\usepackage{xcolor}
\usepackage{longtable,booktabs,array}
\usepackage{calc} 
\usepackage{etoolbox}
\makeatletter
\patchcmd\longtable{\par}{\if@noskipsec\mbox{}\fi\par}{}{}
\makeatother
\IfFileExists{footnotehyper.sty}{\usepackage{footnotehyper}}{\usepackage{footnote}}
\makesavenoteenv{longtable}
\usepackage{graphicx}
\makeatletter
\def\maxwidth{\ifdim\Gin@nat@width>\linewidth\linewidth\else\Gin@nat@width\fi}
\def\maxheight{\ifdim\Gin@nat@height>\textheight\textheight\else\Gin@nat@height\fi}
\makeatother
\setkeys{Gin}{width=1\maxwidth,height=\maxheight,keepaspectratio}
\makeatletter
\def\fps@Figure{htbp}
\makeatother
\usepackage[normalem]{ulem}
\ifLuaTeX
  \usepackage{selnolig}  
\fi
\IfFileExists{bookmark.sty}{\usepackage{bookmark}}{\usepackage{hyperref}}
\IfFileExists{xurl.sty}{\usepackage{xurl}}{} 
\hypersetup{
  hidelinks,
  pdfcreator={LaTeX via pandoc}}

\usepackage{titlesec}

\titleformat{\section}
{\Large\bfseries}{\thesection}{1em}{}

\titleformat{\subsection}
{\small\bfseries}{\thesubsection}{1em}{}

\newcolumntype{Y}{>{\centering\arraybackslash}X}

\usepackage{amsmath,amsfonts}
\usepackage{bm}
\usepackage{array}
\usepackage{textcomp}
\usepackage{stfloats}
\usepackage{url}
\usepackage{verbatim}
\usepackage{graphicx}
\usepackage{color}
\usepackage{tabularx}
\usepackage{booktabs}
\usepackage{multirow}
\usepackage{makecell}
\usepackage{threeparttable}
\usepackage{amssymb}
\usepackage{bbm}
\usepackage{mathtools}
\usepackage{graphicx}%
\usepackage{multirow}%
\usepackage{amsmath,amssymb,amsfonts}%
\usepackage{amsthm}%
\usepackage{mathrsfs}%
\usepackage[title]{appendix}%
\usepackage{xcolor}%
\usepackage{textcomp}%
\usepackage{manyfoot}%
\usepackage{booktabs}%
\usepackage{algorithm}%
\usepackage{algorithmicx}%
\usepackage{algpseudocode}%
\usepackage{listings}%
\usepackage{bbding}
\usepackage{utfsym}
\usepackage{arydshln}
\usepackage{lipsum}
\usepackage{tabularx}

\author{}
\date{}

\begin{document}

\textbf{\LARGE{A Concept-informed Interpretable Artificial Intelligence Framework for Breast Ultrasound Diagnosis}}\\

Shengzhou Zhong$^{1,4,5}$, Ziying Wang$^{1,4,5}$, Liang Li$^{2}$, Lihua Yan$^{2}$, Shuochun Chen$^{2}$, Yudie Zou$^{2}$, Jifan Zhang$^{2}$, Wenhao Li$^{2}$, Huiling Zhang$^{2}$, Jiayin Lu$^{2}$, Ziting Xu$^{2}$, Qianjin Feng$^{1,4,5}$, Zhenwei Shi$^{6,7,*}$, Yan Chen$^{3,*}$, Yingjia Li$^{2,*}$, Yu Zhang$^{1,4,5,*}$, and Zhenyuan Ning$^{1,4,5,8,*}$.

$^{1}$ School of Biomedical Engineering, Southern Medical University, Guangzhou, Guangdong, 510515, China.\\
$^{2}$ Department of Ultrasound, Nanfang Hospital, Southern Medical University, Guangzhou, Guangdong, 510515, China.\\
$^{3}$ Department of Ultrasonic Diagnosis, Zhujiang Hospital, Southern Medical University, Guangzhou, Guangdong, 510282, China.
$^{4}$ Guangdong Provincial Key Laboratory of Medical Image Processing, Southern Medical University, Guangzhou, Guangdong, 510515, China.\\
$^{5}$ Guangdong Province Engineering Laboratory for Medical Imaging and Diagnostic Technology, Southern Medical University, Guangzhou 510515, China.\\
$^{6}$ Department of Radiology, Guangdong Provincial People’s Hospital, Guangdong Academy of Medical Sciences , Southern Medical University, Guangzhou 510080, China.\\
$^{7}$ Guangdong Provincial Key Laboratory of Artificial Intelligence in Medical Image Analysis and Application, Guangzhou 510080, China.\\
$^{8}$ Department of Radiology, Zhujiang Hospital, Southern Medical University, Guangzhou, Guangdong, China

$^{*}$ Corresponding authors: jonnyning@foxmail.com (ZY. Ning), yuzhang@smu.edu.cn (Y. Zhang), lyjia@smu.edu.cn (YJ. Li), smu\_chen@163.com (Y. Chen) and shizhenwei@gdph.org.cn (ZW. Shi).

\section{Abstract}

Most existing artificial intelligence systems for breast ultrasound rely primarily on image-driven prediction, with limited incorporation of the knowledge from the Breast Imaging Reporting and Data System (BI-RADS) and the intermediate assessment processes underlying radiologists' workflow, thereby constraining their interpretability and alignment with clinical practice.
We curated, to our knowledge, the largest multicenter collection of breast ultrasound images paired with structured BI-RADS descriptors, comprising 9,941 patients across 20 independent datasets. 
Across 14 testing datasets, IBUS demonstrated consistent generalization performance across different clinical centers and imaging conditions. 
Incorporating BI-RADS descriptors enabled fine-grained risk stratification, while cross-checking predicted BI-RADS grades against malignancy probabilities identified internally discordant predictions that may warrant additional clinical review. 
In a reader study involving eight radiologists, IBUS assistance increased diagnostic sensitivity among both junior and senior readers, with the potential of guiding radiologists to make targeted revision in risk stratification. 
Overall, IBUS exhibits its potential as a second-stage decision-support system that complements radiologists by providing additional risk estimates and interpretable evidence for diagnostically challenging cases.
The demo video of IBUS platform is publicly available at \textcolor{blue}{https://github.com/zsz1997/IBUS}.

\section{Introduction}

Breast cancer is one of the most common malignancies and the leading cause of cancer-related deaths among women worldwide~\cite{heer2020global}.
In 2024, approximately 2.4 million women were newly diagnosed, with an estimated 694,000 deaths globally~\cite{sung2026global}.
Many studies have demonstrated that early diagnosis and timely treatment are strongly associated with improved clinical outcome~\cite{goldhirsch2013personalizing,xiong2025breast}.
Currently, mammography is considered as the standard screening tool, but its sensitivity is markedly reduced in women with dense breast tissue~\cite{harada2021evaluation,gilbert2025comparison}, and its accessibility is limited in sources-constrained settings~\cite{qian2021prospective,luo2024deep}.
In this context, breast ultrasound (BUS) has emerged as an important complementary imaging modality due to its non-invasiveness, accessibility, and effectiveness in dense breasts~\cite{dan2024diagnostic,chang2025hallmarks}.
Therefore, BUS plays an increasingly important role in early diagnosis.

To standardize ultrasound assessment and facilitate clinical decision, radiologists rely on the Breast Imaging Reporting and Data System (BI-RADS)~\cite{atlas2013breast} to characterize the breast lesion using standardized descriptors (including shape, margin, orientation, echo pattern, posterior features, calcifications, and so on) and evaluate its malignancy likelihood.
Despite its widespread adoption, BI-RADS assessment remains inherently challenging (especially for junior radiologists with less experience) due to speckle noise, low contrast, and ambiguous lesion boundary in ultrasound images, leading to substantial inter-observer variability~\cite{shen2021artificial}.
Such variability might result in misdiagnosis of malignant lesions or unnecessary biopsies for benign cases~\cite{myers2015benefits,qian2025multimodal}.
In recent years, deep learning has demonstrated a remarkable ability to automatically learn discriminative representations from imaging data~\cite{lecun2015deep,bi2019artificial,mckinney2020international} and has been increasingly applied to assist BUS diagnosis~\cite{jiang2017artificial,leibig2022combining}.
However, despite their promising performance, most studies still rely primarily on image-driven representation learning and fail to incorporate the clinical prior from BI-RADS descriptors to align with established diagnostic criteria, thereby limiting their interpretability and clinical trustworthiness~\cite{zhou2023transformer,yao2023virtual}.

\begin{figure}[!t]
	\centering
	\includegraphics[width=1\textwidth]{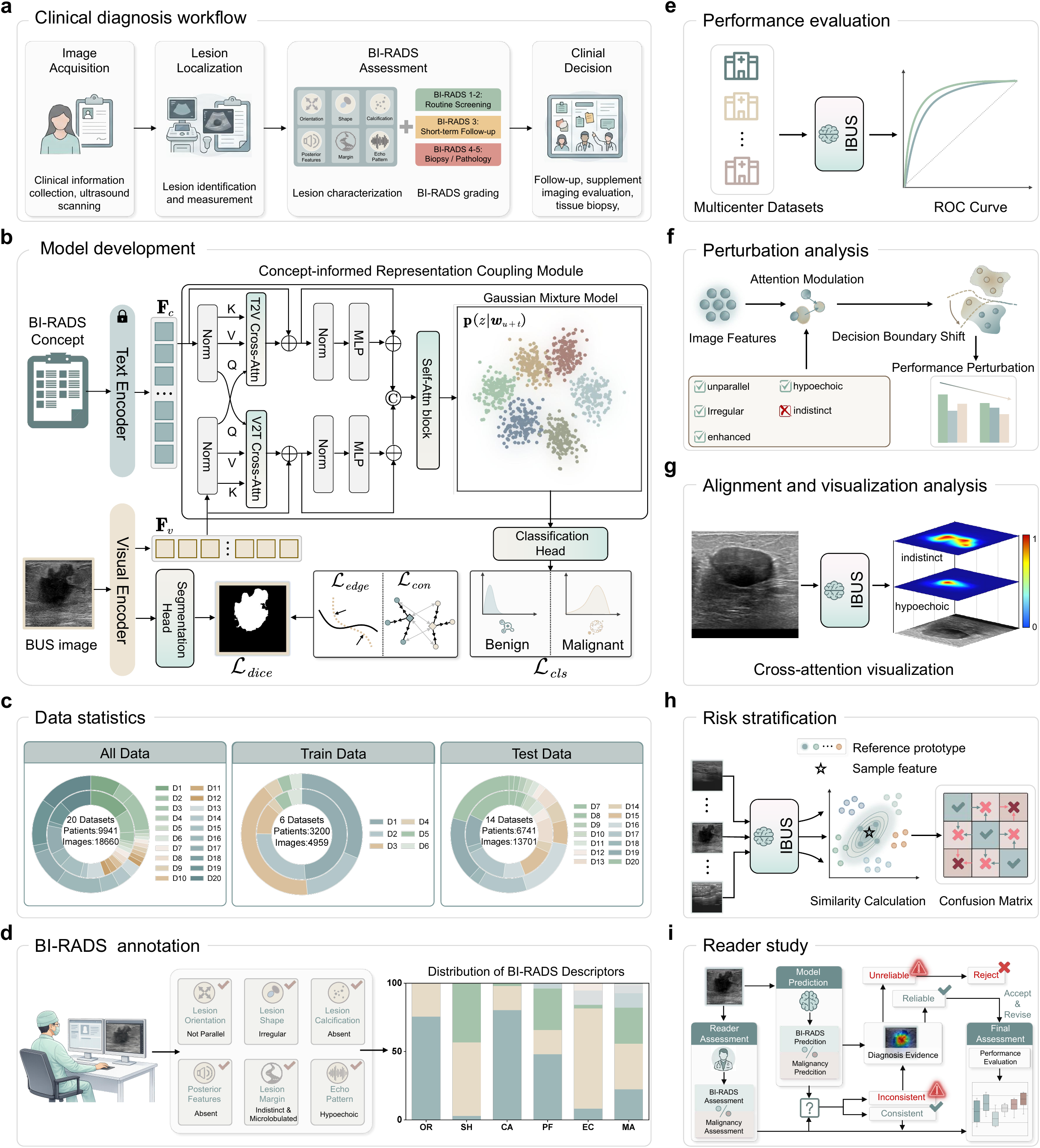}
	\caption{\textbf{Overview of the dataset and study design.} 
        \textbf{a}, Workflow of clinical diagnosis under BI-RADS guideline. The workflow mainly consists of image acquisition, lesion localization, BI-RADS assessment, and clinical decision.
        \textbf{b}, Architecture of the proposed IBUS framework. IBUS mainly consists of four key components, including a dual-stream feature extractor, a segmentation head, a concept-informed representation coupling module and a classification head.
		\textbf{c}, Data statistics for all, training and testing data. The study cohort comprised 20 independent datasets (i.e., D1\textasciitilde D20) from multiple centers, in which 6 datasets (i.e., D1\textasciitilde D6) and 14 datasets (i.e., D7\textasciitilde D20) were used to develop and evaluate IBUS.
		\textbf{d}, Annotation of structured BI-RADS descriptors for ultrasound images.
        \textbf{e}, Performance evaluation on multicenter datasets.
        \textbf{f}, Perturbation analysis on different BI-RADS descriptors.
        \textbf{g}, Semantic-visual alignment and visualization analysis on model's attention.
        \textbf{h}, Risk stratification based on sample similarity retrieval.
        \textbf{i}, Reader study for assessing IBUS's capability in assisting clinicians evaluation.}\label{overview}
\end{figure}

As another evident limitation, current models generally ignore the essential components of realistic clinical BI-RADS assessment workflow.
Clinically, BI-RADS assessment follows a structured workflow that progresses from lesion localization to risk stratification (as shown in \textbf{Fig}.~\ref{overview}a).
First, radiologists manually identify and measure suspicious lesions to characterize their morphological features and imaging patterns.
Subsequently, these characterized findings are interpreted in conjunction with relevant clinical information to assign a BI-RADS category, which provides an estimate of malignancy risk and informs subsequent patient triage and disease management, including routine follow-up, additional imaging evaluation, or tissue biopsy~\cite{atlas2013breast}.
Such a workflow provides transparent and traceable diagnostic evidence that underpins clinical decision-making. 
However, most existing AI systems bypass this workflow and instead formulate a binary diagnosis task (i.e., malignant vs. benign), which remains weakly aligned with the intermediate reasoning process of radiologists and limits their deployment in real clinical scenario~\cite{jiao2024usfm,jiang2025pretraining,yu2026foundation}.

In this work, we present IBUS, a concept-informed interpretable artificial intelligence framework for breast ultrasound diagnosis.
Unlike previous studies that primarily pursue incremental performance improvements, we focus on two key aspects rather than merely demonstrating technical superiority.
First, IBUS improves image representation learning and model interpretability by incorporating BI-RADS descriptors into the training process.
Second, IBUS mimics the BI-RADS assessment workflow by explicitly modeling lesion location, feature interpretability and risk stratification, which provides transparent and traceable evidence that aligns with clinical reasoning and facilitates clinical deployment.
To develop and evaluate IBUS, we curated the largest breast ultrasound dataset with paired ultrasound images and BI-RADS descriptors, including 9,941 participants of 20 datasets from multiple medical centers. 
The experimental results demonstrate IBUS achieves promising generalization and robustness on multiple external datasets. 
IBUS effectively bridges visual representations (derived from ultrasound images) and semantic features (encoded in BI-RADS descriptors) and identifies the lesion margin as the most influential descriptor for breast ultrasound diagnosis.
Meanwhile, IBUS closely aligns with radiologists in risk stratification, and joint consideration of BI-RADS grade and malignancy probabilities enables identification of discordant cases that may warrant additional clinical review. 
More importantly, IBUS assistance improved radiologists' diagnostic performance in malignant lesions, with the potential of guiding radiologists to make targeted changes in risk stratification when dealing with uncertain cases.
Rather than replacing radiologists, IBUS is designed to assist clinical decision-making by generating transparent and interpretable evidence alongside malignancy risk estimates, thereby supporting radiologists throughout the ultrasound assessment workflow.

\section{Results}\label{sec2}

\begin{figure}[!t]
	\centering
	\includegraphics[width=1\textwidth]{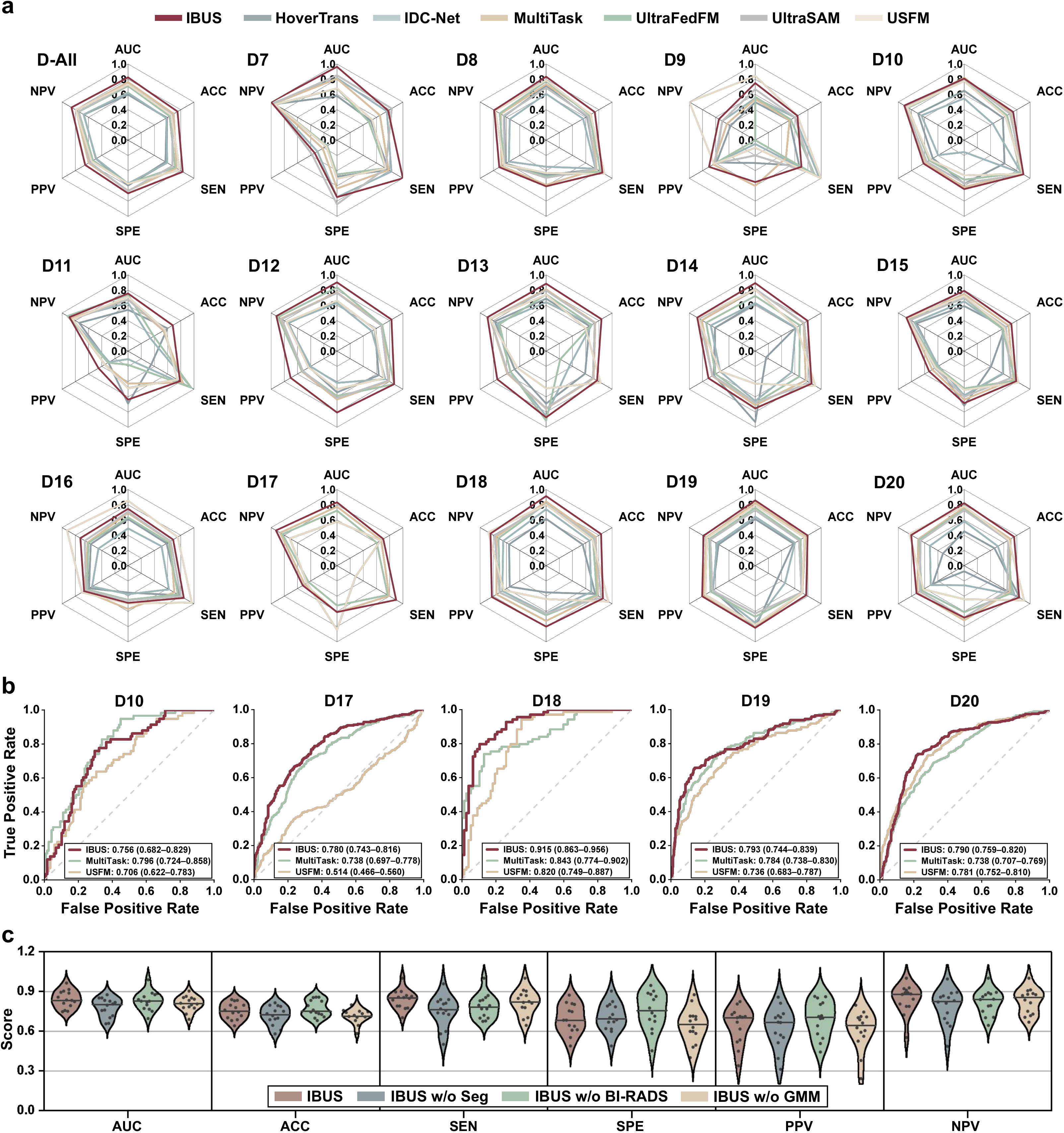}
	\caption{\textbf{Model generalization across multicenter datasets.}
	\textbf{a}, Radar plot for diagnosis metrics across six competing methods on all (D-All) and each (D7-D20) test sets, with six metrics of AUC, ACC, SEN, SPE, PPV, and NPV compared.
	\textbf{b}, ROC curves on diagnostically challenging cases (BI-RADS 3 and 4 lesions) across Top-2 models (i.e., MultiTask and USFM) on five datasets (i.e., D10, D17\textasciitilde D20).
	\textbf{c}, Ablation study on three variants: (1) IBUS w/o Seg: removing the segmentation head; (2) IBUS w/o BI-RADS: removing the stream tailored for BI-RADS descriptors; (3) IBUS w/o GMM: removing risk stratification modeling.
	}\label{comparison_performance}
\end{figure}

\begin{figure}[!h]
	\centering
	\includegraphics[width=1\textwidth]{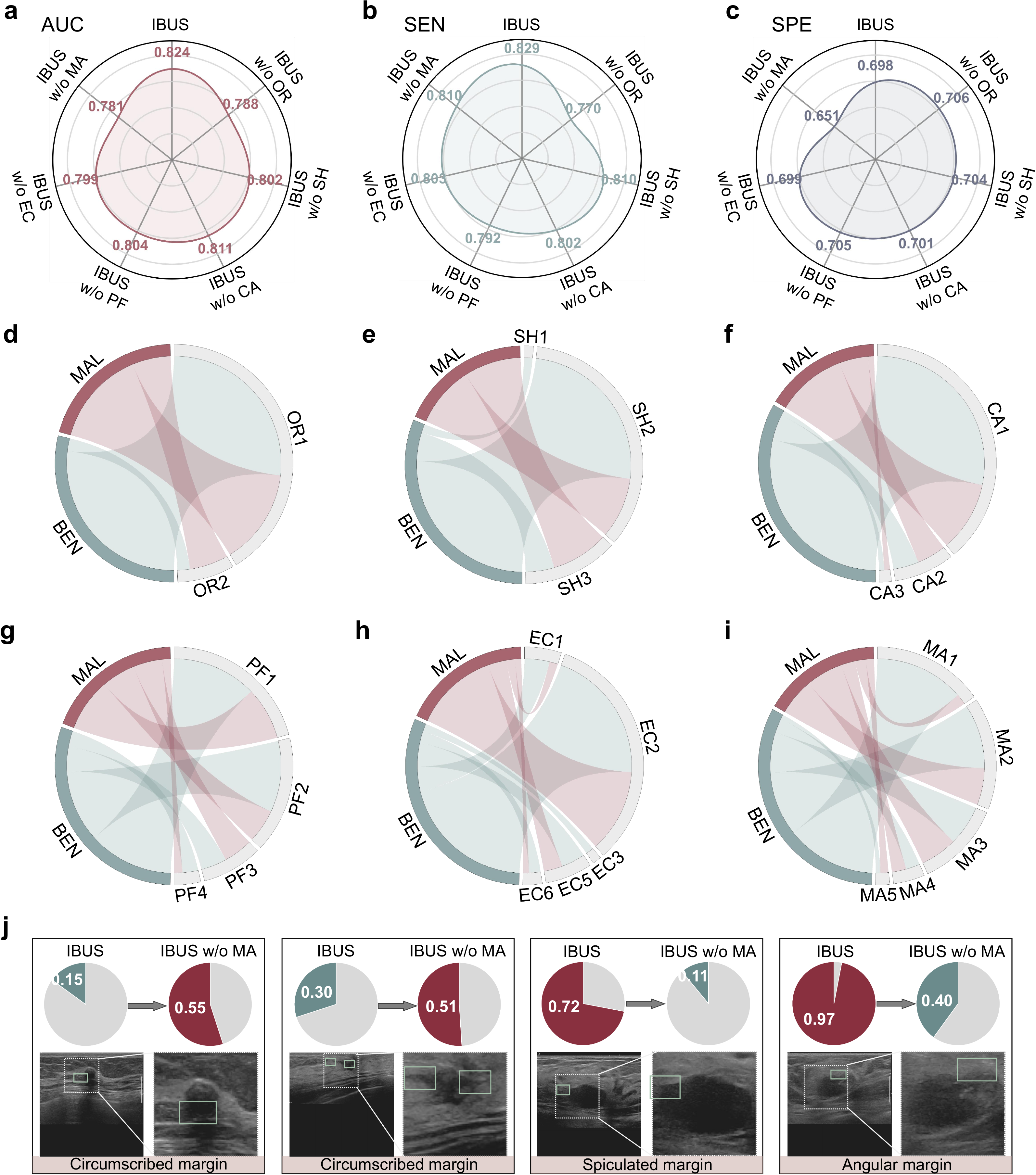}
	\caption{\textbf{Perturbation analysis of BI-RADS descriptors.}
		\textbf{a}-\textbf{c}, Comparison of prediction performance across IBUS and its variants after removing individual BI-RADS descriptor groups, with three metrics of AUC, SEN, and SPE.
		\textbf{d}-\textbf{i}, Distribution of descriptors among misclassified samples from IBUS w/o OR (\textbf{d}), IBUS w/o SH (\textbf{e}), IBUS w/o CA (\textbf{f}), IBUS w/o PF (\textbf{g}), IBUS w/o EC (\textbf{h}) and IBUS w/o MA (\textbf{i}), respectively. ``MAL'' and ``BEN'' denote malignant and benign, respectively.
        \textbf{j}, Four representative misclassified cases from IBUS w/o MA.
	}\label{text_influence}
\end{figure}

\subsection{The overview of study design}

The overall study design is summarized in {Figure}~\ref{overview}.
To align with the clinical workflow, we developed IBUS, a concept-informed interpretable artificial intelligence framework that integrates ultrasound images with structured BI-RADS descriptors for breast ultrasound diagnosis.
{Figure}~\ref{overview}b exhibits the architecture of IBUS that mainly consists of four key components, including a dual-stream feature extractor, a segmentation head, a concept-informed representation coupling module and a classification head. 
Specifically, the dual-stream feature extractor consisting of a pretrained ResNet-34~\cite{he2016deep} as visual encoder and a frozen CLIP as text encoder~\cite{radford2021learning} to jointly extract visual and textual features from ultrasound images and BI-RADS descriptors, respectively.
Subsequently, the segmentation head is constructed to locate breast lesion, which is constrained by boundary-aware semantic contrastive learning.
And the concept-informed representation coupling module is designed to incorporate BI-RADS descriptors into the network, in which visual and textual features are integrated through cross-attention and self-attention blocks, followed by a Gaussian mixture model (GMM) for risk stratification (i.e., BI-RADS grading). 
Finally, the classification head is used for breast cancer diagnosis.
As shown in {Figure}~\ref{overview}c and Supplementary Table~1, the study cohort comprised 20 independent datasets (denoted as D1\textasciitilde D20) collected from different clinical centers, including 9,941 patients and 18,660 ultrasound images.
And 6 datasets (i.e., D1\textasciitilde D6, 3,200 patients with 4,959 images) and 14 datasets (i.e., D7\textasciitilde D20, 6,741 patients with 13701 images) were used to develop and evaluate IBUS.
All BI-RADS descriptors were retrospectively annotated by radiologists according to the BI-RADS ultrasound lexicon, including lesion shape (SH), orientation (OR), calcifications (CA), posterior features (PF), echo pattern (EC), and margin (MA)(as illustrated in {Figure}~\ref{overview}d and Supplementary Table~2).
We conducted comprehensive experiments to assess IBUS from five aspects.
For performance evaluation, we compared IBUS with state-of-the-art ultrasound diagnosis methods on 14 external testing datasets ({Figure}~\ref{overview}e).
Perturbation analysis was performed by individually removing each BI-RADS descriptors to quantify its contribution to model performance ({Figure}~\ref{overview}f).
Subsequently, explicit links between the image representations learned by IBUS and BI-RADS descriptors were established through semantic–visual alignment ({Figure}~\ref{overview}g).
Beyond conventional binary diagnosis, we further evaluated IBUS for risk stratification task.
Finally, we conducted reader study to evaluate the clinical utility of IBUS in assisting radiologists with varying levels of experience in breast ultrasound assessment ({Figure}~\ref{overview}i).
Please see \textbf{Code availability} for the demo video of IBUS platform.

\begin{figure}[!h]
	\centering
	\includegraphics[width=1\textwidth]{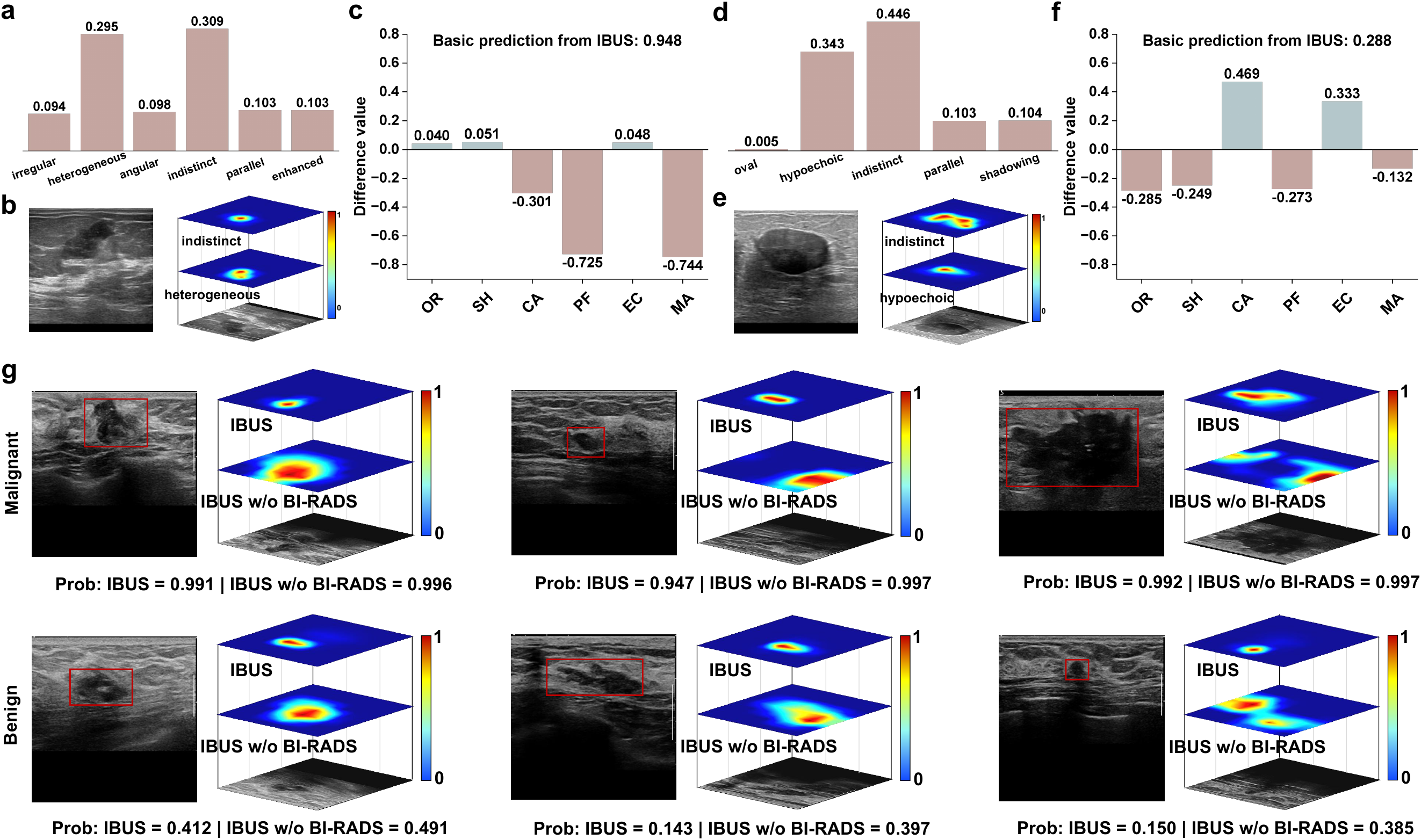}
	\caption{\textbf{Semantic--visual alignment and visualization analysis.}
		\textbf{a}-\textbf{c}, Semantic--visual alignment and cross-attention visualization on one representative malignant case.
        \textbf{a}, Visualization of attention weights across different BI-RADS descriptors.
        \textbf{b}, cross-attention maps across important BI-RADS descriptors.
        \textbf{c}, The value change of malignancy probability after removing different BI-RADS descriptors.
        \textbf{d}-\textbf{f}, Semantic--visual alignment and cross-attention visualization on one representative benign case.
        \textbf{d}, Visualization of attention weights across different BI-RADS descriptors.
        \textbf{e}, Cross-attention maps across important BI-RADS descriptors.
        \textbf{f}, The value change of malignancy probability after removing different BI-RADS descriptors.
        \textbf{g}, Class activation maps (CAM) visualization from IBUS and IBUS w/o BI-RADS, respectively.
	}\label{visualization}
\end{figure}

\begin{figure}[!h]
	\centering
	\includegraphics[width=1\textwidth]{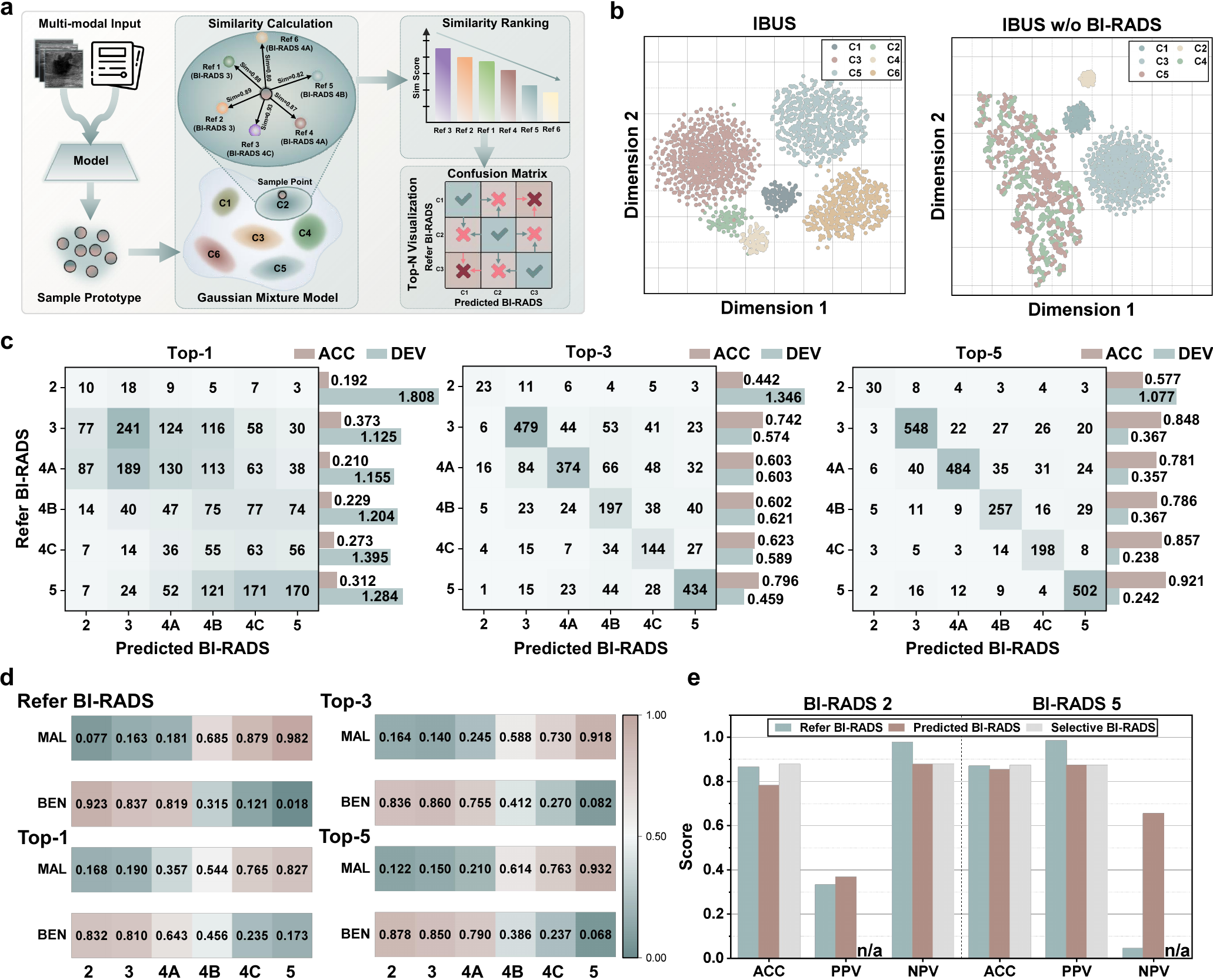}
	\caption{\textbf{BI-RADS risk stratification performance.}
		\textbf{a}, Process of risk stratification evaluation.
        \textbf{b}, T-SNE maps visualization of features from IBUS and IBUS w/o BI-RADS.
		\textbf{c}, Top-$N$ confusion matrices for BI-RADS risk stratification between reference BI-RADS and IBUS-predicted BI-RADS. The retrieval accuracy (i.e., ACC) and deviation (i.e., DEV) score for each BI-RADS grade are displayed in the bar chart on the right side of the confusion matrices.
		\textbf{d}, Proportion of benign and malignant cases in six categories derived from reference BI-RADS grades and Top-$1$\textasciitilde Top-$5$ IBUS-predicted BI-RADS grades.
        \textbf{e}, Comparison of IBUS's diagnostic performance on BI-RADS 2 and 5 based on reference BI-RADS grades, Top-$1$ IBUS-predicted BI-RADS grades, and selective Top-$1$ IBUS-predicted BI-RADS grades, respectively. ``n/a'' means no corresponding class can be used to calculate the metric.
	}\label{risk_stra}
\end{figure}

\subsection{Model generalization across multicenter datasets}
To comprehensively evaluate model's generalization, we compared IBUS with six state-of-the-art (SOTA) ultrasound classification methods, including two single-task models (i.e., IDC-Net~\cite{yi2024idc} and HoverTrans~\cite{mo2023hover}), a multi-task learning framework (i.e., MultiTask~\cite{aumente2025multi}), and three ultrasound foundation/pre-trained models (i.e., UltraFedFM~\cite{jiang2025pretraining}, UltraSAM~\cite{meyer2025ultrasam}, and USFM~\cite{jiao2024usfm}).
All compared models were reproduced based on their paper and trained/fine-tuned with same training data.
Fig.~\ref{comparison_performance}a exhibits the patient-level performance (i.e., average probability of all images from patient) of all competing methods across entire test sets (denoted as D-All) and 14 independent testing datasets (D7-D20).
Across the multicenter testing datasets, IBUS achieved the best performance on 11 of 14 datasets and yielded the highest overall performance (D-All: AUC, 0.824; 95\% CI, 0.813–0.834; \(p<0.001\)), demonstrating robust generalization across diverse clinical settings.
Given the clinical significance of BI-RADS 3 and 4 (including 4A, 4B and 4C), we further compared IBUS primarily with the two strongest models (i.e., MultiTask and USFM) across five datasets (i.e., D10 and D17\textasciitilde D20). 
As shown in Fig.~\ref{comparison_performance}b, IBUS achieved the highest AUROC on 4/5 datasets, indicating promising potential in distinguishing diagnostically challenging lesions.
Furthermore, we conducted ablation study to validate the efficacy of key components by constructing different variants, including 
(1) IBUS w/o Seg: removing the segmentation head; (2) IBUS w/o BI-RADS: removing the stream tailored for BI-RADS descriptors; (3) IBUS w/o GMM: removing risk stratification modeling.
As illustrated in Fig.~\ref{comparison_performance}c, all variants exhibited performance degradation compared with the complete IBUS. These results support the benefits of mimicking the clinical ultrasound assessment workflow and incorporating BI-RADS descriptors into the model.

\subsection{Perturbation analysis of BI-RADS descriptors}
To quantify the contribution of all BI-RADS descriptors, we conducted perturbation analysis by removing each descriptor group to retrain IBUS.
Fig.~\ref{text_influence}a illustrates the ROC curves.
The overall AUC consistently decreased relative to the complete IBUS when any group of BI-RADS descriptors was removed, with the largest decrease of 5.4\% observed for IBUS w/o MA (AUC, 0.781), indicating that different BI-RADS descriptors provide complementary diagnostic evidences.
Fig.~\ref{text_influence}b-c further illustrates the sensitivity and specificity results.
A significant decrease in the sensitivity was observed when any group of BI-RADS descriptors was removed (\(p<0.01\)), whereas the specificity remained largely unchanged, with no significant differences observed except for IBUS w/o MA (\(p<0.001\)), which suggests that the margin descriptors may serve as a key imaging determinant for breast cancer diagnosis.
Furthermore, we summarized the BI-RADS descriptor distributions among misclassified cases, in illustrated in Fig.~\ref{text_influence}d-i.
The removal of all descriptors resulted in the misclassification of benign and malignant lesions, with a higher proportion of errors observed among benign lesions. 
Misclassified lesions following orientation/shape/calcifications/echo pattern removal were predominantly characterized by OR1/SH2/CA1/EC2, whereas those following posterior features/margin removal were more evenly distributed across PF1\textasciitilde PF3/MA1\textasciitilde MA4.
Finally, Fig.~\ref{text_influence}i shows some representative cases for IBUS \textit{v.s.} IBUS w/o MA, with others in Appendix Fig.~S4.
Removing margin descriptors substantially increased confusion between benign lesions with circumscribed margins and malignant lesions with indistinct or spiculated/angular margins.
These observations further demonstrate the core contribution of BI-RADS descriptors to robust lesion characterization.

\subsection{Semantic--visual alignment and visualization analysis}

Here, we first concentrate on the alignment between BI-RADS descriptors and image regions using two representative cases, including a malignant invasive ductal carcinoma and a benign fibroadenoma with lactational adenoma.
Fig.~\ref{visualization}a and d show the descriptor-level importance scores.
For the malignant case, the model assigned the highest weights to ``indistinct'' and ``heterogeneous,'' whereas ``indistinct'' and ``hypoechoic'' received greater attention in the benign one.
These findings suggest that the model selectively emphasizes diagnostically relevant BI-RADS descriptors rather than assigning uniform weights to all inputs.
Fig.~\ref{visualization}b and e present the cross-attention maps corresponding to Top-2 important descriptors.
We can observe that high attention associated with ``indistinct'' was concentrated around blurred lesion margins, whereas ``heterogeneous'' and ``hypoechoic'' mainly attended to the internal regions.
This demonstrates that IBUS enables effective alignment between BI-RADS descriptors and image regions.
Fig.~\ref{visualization}c and f illustrate the changes in predicted malignancy probability after removing individual BI-RADS descriptors for two individual cases.
Removing individual BI-RADS descriptors affected the model’s ability to recognize subtle imaging features. 
For example, in the malignant case, removing the ``indistinct'' descriptor reduced the model's attention to suspicious lesion margins and was accompanied by a 0.744 decrease in the predicted malignancy probability.
Conversely, since the presence of internal echogenic variations made the ``hypoechoic'' descriptor important for correct assessment on the benign case, removing this descriptor increased the predicted malignancy probability by 0.333, leading to an incorrect prediction. 
Finally, Fig.~\ref{visualization}g illustrates the class activation maps (CAMs) generated by IBUS and IBUS w/o BI-RADS.
With BI-RADS descriptors, IBUS produced more compact activation maps that better overlapped with lesion regions, whereas IBUS w/o BI-RADS easily attended to irrelevant structures despite yielding correct predictions.
Thus, incorporating BI-RADS descriptors improves not only predictive performance but also model interpretability.

\subsection{Model performance on BI-RADS risk stratification}

BI-RADS risk stratification plays a key role in breast cancer diagnosis and subsequent clinical management.
Based on the mixture representations learned by the GMM, we performed similarity-based risk stratification for BI-RADS grading (see \textbf{Methods} for details), as illustrated in Fig.~\ref{risk_stra}a. 
Each sample was first assigned to its most likely Gaussian component, defining a corresponding risk stratum. 
Feature similarities were then computed between the query sample and retrieved reference cases within the same component. 
Finally, the proportion of all BI-RADS grades in Top-$N$ retrieved cases were calculated and ranked, then those with higher proportion were derived as candidate risk grades.
We first visualized the embeddings from GMM using t-SNE (Fig.~\ref{risk_stra}b).
Compared with IBUS w/o BI-RADS, the representations from IBUS exhibit clearer inter-cluster separation and greater intra-cluster compactness.
Such more structured embedding distribution suggests that incorporating BI-RADS descriptors helps organize lesions into more distinguishable latent clusters.
Fig.~\ref{risk_stra}c shows the confusion matrices under the Top-$1$, Top-$3$, and Top-$5$ evaluation criteria and the corresponding retrieval accuracy (i.e., ACC) and deviation (i.e., DEV) score for each BI-RADS category.
As the retrieval range expanded, mean retrieval accuracy increased from 0.285 (0.192--0.373) for Top-$1$ to 0.834 (0.577--0.921) for Top-$5$, accompanied by a reduction in the deviation score from 1.191 (1.095--1.808) to 0.339 (0.242\textasciitilde 1.077).
Apparently, matching errors were largely concentrated around neighboring risk categories, for example, BI-RADS 4A lesions were frequently matched to BI-RADS 3 or 4B.
We also calculated the proportions of benign and malignant lesions within each BI-RADS category, as shown in Fig.~\ref{risk_stra}d. 
Similar to reference BI-RADS grades (Refer BI-RADS), IBUS-predicted grades showed a progressive increase in the proportion of malignant lesions and a corresponding decrease in benign lesions with increasing BI-RADS category. 
Such pattern indicates that IBUS-predicted grades captured a clinically consistent risk gradient.
Notably, performance remained relatively less satisfactory at the BI-RADS 2 and 5 on Top-$1$ evaluation criteria.
We therefore further summarized diagnostic performance within these two categories according to the reference and IBUS-predicted BI-RADS grades (Fig.~\ref{risk_stra}e). 
To determine whether inconsistencies between risk stratification and malignancy prediction could identify potentially unreliable predictions, we additionally defined a selective analysis (Selective BI-RADS) that excluded discordant cases, namely lesions predicted as BI-RADS 2 with a malignancy probability \(>0.5\) and those predicted as BI-RADS 5 with a malignancy probability \(<0.5\). 
Across most metrics, diagnostic performance stratified by IBUS-predicted grades was comparable to or even better than that based on reference BI-RADS grades. 
After excluding discordant cases, no malignant cases remained in the predicted BI-RADS 2 group and no benign cases remained in the predicted BI-RADS 5 group, rendering the corresponding PPV or NPV undefined (``n/a''). 
Obviously, cross-checking predicted BI-RADS grades against malignancy probabilities may provide a simple mechanism for flagging internally discordant predictions which might need additional clinical review.

\subsection{Reader study}

\begin{figure}[!h]
	\centering
	\includegraphics[width=0.98\textwidth]{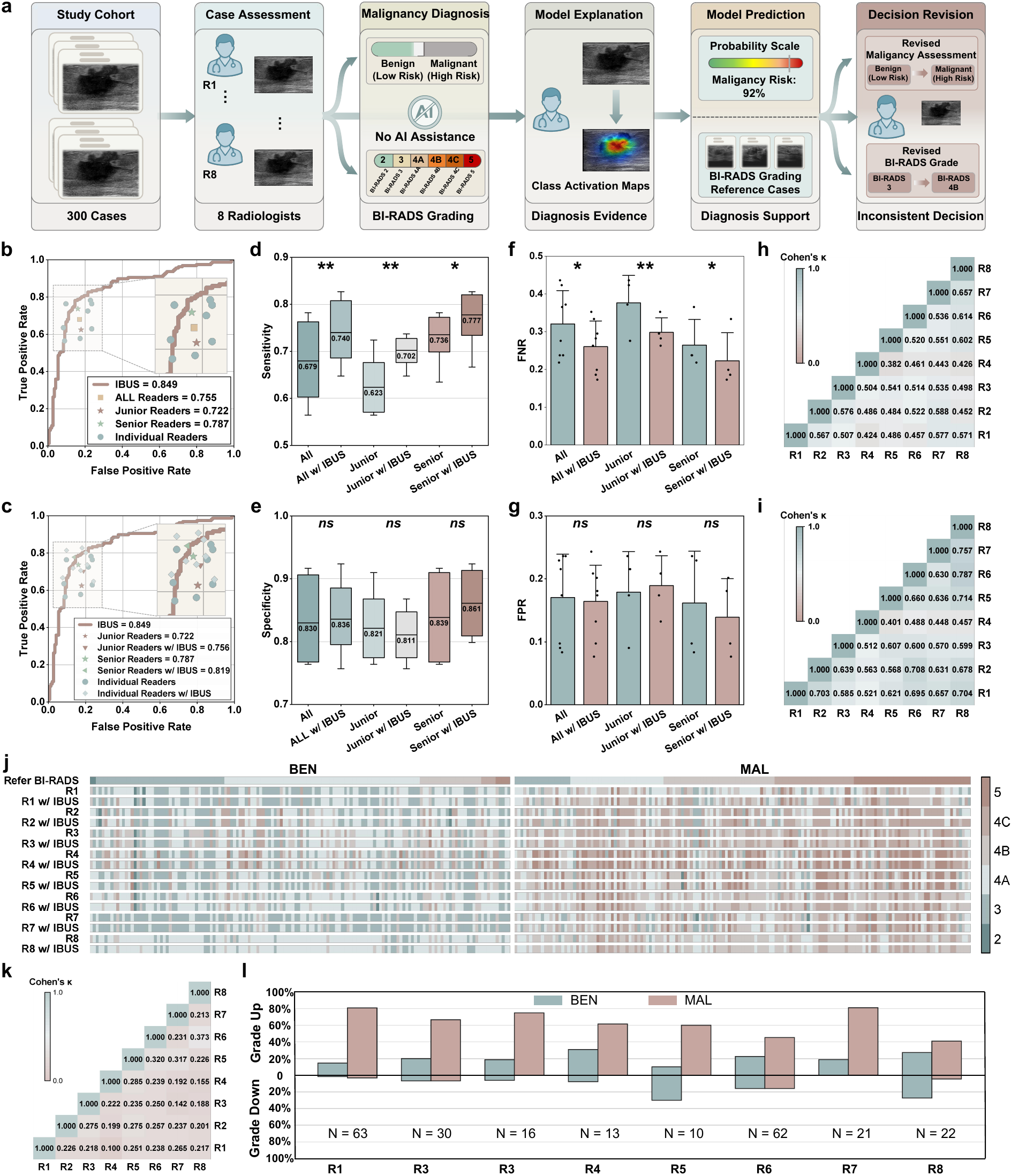}
	\caption{\textbf{AI-assisted diagnosis and risk stratification performance in breast lesion assessment.}
		\textbf{a}, The flowchart of reader study.
		\textbf{b}, ROC curve from IBUS, all readers (i.e., R1\textasciitilde R8), junior readers (i.e., R1\textasciitilde R4), senior readers (i.e., R5\textasciitilde R8) and individual readers without IBUS assistance.
		\textbf{c}, ROC curve from IBUS, all readers, junior readers, senior readers and individual readers without and with IBUS assistance.
		\textbf{d}-\textbf{e}, Comparison of all readers, junior readers and senior readers without and with IBUS assistance on sensitivity (\textbf{d}) and specificity (\textbf{e}).
		\textbf{f}-\textbf{g}, Comparison of all readers, junior readers and senior readers without and with IBUS assistance on FNR (\textbf{f}) and FPR (\textbf{g}).
		\textbf{h}-\textbf{i}, Cohen's kappa values on malignancy assessment between readers without (\textbf{h}) and with IBUS assistance (\textbf{i}).
		\textbf{j}, BI-RADS grades of benign and malignant cases from eight readers with and without IBUS assitance.
		\textbf{k}, Cohen's kappa values on risk stratification between all readers without IBUS assistance.
		\textbf{l}, Statistics of benign and malignant cases with risk stratification modifications. 'Grade Up' and 'Grade Down' refer to raising or lowering the BI-RADS grade, respectively.
	}\label{ai_assisted}
\end{figure}

Fig.~\ref{ai_assisted}a illustrates the flowchart of reader study (More details refer to \textbf{Methods} with IBUS demo video at \textcolor{blue}{https://github.com/zsz1997/IBUS}).
Here, eight radiologists evaluated 300 cases randomly selected from our private institutional cohorts in two stages. 
First, all readers independently assessed each case for BI-RADS risk stratification and malignancy diagnosis. 
After a more than two-week washout period, they were provided with IBUS predictions, including the Top-$3$ predicted BI-RADS grades and malignancy probability based on the input BI-RADS descriptors from each reader, together with corresponding interpretable evidence (i.e., CAMs). 
The readers then reassessed each case with IBUS assistance and provided BI-RADS grades and lesion diagnosis.
Firstly, we calculated the classification performance of IBUS on the study cohort.
The ROC curves and quantitative results are shown in Fig.~\ref{ai_assisted}b-c and Table~S3 in the Appendix.
Overall, IBUS achieved an AUC of 0.849 (95\% CI: 0.802–0.892) and outperformed both junior (i.e., R1\textasciitilde R4, AUC: 0.722, \textit{p} < 0.05) and senior radiologists (i.e., R5\textasciitilde R8, AUC: 0.787, \textit{p} < 0.05) (Fig.~\ref{ai_assisted}b).
Compared with eight radiologists, IBUS exhibited significantly higher sensitivity (0.846 \textit{vs}.\ 0.679; \textit{p} < 0.05).
Radiologists achieved sensitivity ranging from 0.564 to 0.782 and higher specificity from 0.764 to 0.917.
To assess the practical value of IBUS in clinical workflows, the same eight radiologists were invited to re-evaluate all cases with IBUS assistance.
As shown in Fig.~\ref{ai_assisted}d-e and Table~S3 in Appendix, IBUS assistance relatively improved the average sensitivity by 8.8\% and specificity by 0.7\%.
Notably, junior radiologists R1 and R2 showed marked improvements, with sensitivity increasing from 0.564 and 0.577 to 0.718 and 0.705, respectively.
AI assistance significantly reduced false-negative rate (FNR), as shown in Fig.~\ref{ai_assisted}f–g.
The FNR decreased from 32.1\% (range: 21.8\%–43.6\%) to 26.0\% (range: 17.3\%–35.3\%; \textit{p} < 0.05), with a reduction in false-negative diagnoses.
Such results indicate that IBUS provides effective decision support for readers varying experience levels, with particularly pronounced benefit in malignant lesion diagnosis.
Finally, inter-reader agreements among all readers are also improved, with the mean Cohen's kappa increasing from 0.517 (range: 0.382–0.657) to 0.612 (range: 0.401–0.787) (as shown in Fig.~\ref{ai_assisted}h–i).


We further evaluated changes in radiologists' BI-RADS risk stratification with IBUS assistance (Figures~S5 and S6 in Appendix; Fig.~\ref{ai_assisted}j--l). 
Across readers, most lesions were assigned BI-RADS grades between 3 and 4C, and grading discrepancies occurred predominantly between adjacent categories (Fig.~S5 in Appendix). 
For example, reference BI-RADS 4A lesions were frequently assigned as BI-RADS 3 or 4B. Agreement with the reference grades generally increased with reader experience, suggesting that fine-grained BI-RADS assessment remained dependent on clinical experience. 
During independent assessment, inter-reader agreement was generally low, with a mean kappa of 0.234 (0.100\textasciitilde 0.373) (see in Fig.~\ref{ai_assisted}k), indicating substantial variability in BI-RADS grading across readers. 
Stratified analyses of benign and malignant lesions further showed that both IBUS and most readers predominantly assigned benign lesions to lower-risk categories (BI-RADS 3/4A) and malignant lesions to higher-risk categories (BI-RADS 4B\textasciitilde 5) (Fig.~\ref{ai_assisted}j). 
Reader assignments remained largely consistent between without and with IBUS assistance, with a mean paired kappa of 0.893 (0.725\textasciitilde 0.961) in benign cases assessment and 0.812 (0.553\textasciitilde 0.948) in malignant one (Fig.~S6 in Appendix), leaving most initial assessments unchanged while a subset of cases was reclassified. 
As shown in Fig.~\ref{ai_assisted}l, the predominant change was upgrading of malignant lesions, accounting for 73.8\% (175/237) of all modified cases in assessment, whereas malignant lesions were downgraded in only 6.3\% (15/237). 
Changes in benign lesions were less frequent, with 10.1\% (24/237) downgraded and 19.8\% (47/237) upgraded. 
The tendency to upgrade malignant lesions was observed in both junior and senior readers, accounting for 91 and 60 modified assessments, respectively. 
Grade changes in the opposite direction were relatively uncommon for malignant lesions and occurred predominantly within intermediate BI-RADS categories (i.e., 4A--4C), where fine-grained risk differentiation remains challenging.
Together with the high agreement between independent and IBUS-assisted BI-RADS grading, these results indicate that IBUS assistance left most initial assessments unchanged but was associated with directional risk reclassification in a subset of cases, most notably toward higher BI-RADS categories for malignant lesions.

\section{Discussion}

In this large-scale multicenter study involving 9,941 patients across 20 independent datasets with paired BI-RADS descriptors, we developed and validated IBUS, a concept-informed artificial intelligence framework that integrates structured BI-RADS descriptors into breast ultrasound analysis. 
In contrast to conventional AI systems that primarily formulate breast lesion assessment as a binary classification task, IBUS more closely mirrors the clinical BI-RADS workflow by embedding established BI-RADS reasoning structures into breast ultrasound AI.
Across diverse clinical centers and imaging conditions, IBUS demonstrated multicenter generalizability, provided clinically meaningful interpretations and risk stratification, and effectively assisted radiologists in diagnostic decision-making. 
Collectively, integrating standardized radiological knowledge from BI-RADS may provide a promising pathway toward breast ultrasound AI that is interpretable, generalizable and more closely aligned with clinical reasoning.

One of the major barriers to the clinical translation of breast ultrasound AI is the limited generalizability~\cite{shen2021artificial}. 
Ultrasound images exhibit substantial variability owing to differences in operators, acquisition protocols, and scanner manufacturers, resulting in pronounced domain shifts across institutions~\cite{vega2025overcoming,zhao2026clinic}. 
Previous studies have shown that incorporating structured clinical information can substantially improve model robustness compared with image-only approaches~\cite{Pfob2022,qian2025multimodal,Luo2025}. 
Consistent with these observations, IBUS achieved an overall AUC of 0.826 across 14 independent datasets and consistently outperformed CNN-based and Transformer-based methods~\cite{yi2024idc,mo2023hover,aumente2025multi}. 
Notably, several competing foundation models~\cite{jiang2025pretraining,jiao2024usfm,meyer2025ultrasam} also achieved promising performance on internal validation datasets, but their performance deteriorated considerably across external datasets.
This suggest that purely image-driven representations may remain sensitive to site-specific characteristics on breast ultrasound diagnosis.
In contrast, BI-RADS descriptors provide standardized semantic abstractions established through decades of clinical consensus and may offer more clinically consistent representations across heterogeneous imaging conditions.
By incorporating the structured clinical knowledge into representation learning, IBUS introduces clinically meaningful inductive biases that are less dependent on scanner characteristics or imaging protocols, which supports the growing consensus that clinically grounded visual representations can provide more transferable and reliable features for real-world breast ultrasound AI systems~\cite{Pfob2022,Ali2025}.

A key motivation for IBUS was to narrow the gap between conventional AI predictions and the structured assessment process used in clinical breast ultrasound. 
In routine practice, instead of making binary diagnostic decisions, radiologists characterize diagnostically relevant imaging features and integrate these findings with other available information to assign a BI-RADS category that informs subsequent management~\cite{atlas2013breast}. 
Although many approaches have increasingly incorporated textual information into breast imaging models~\cite{li2024ultrasound,alsharid2025public,le2026u2}, the contribution of individual clinical descriptors to model decisions and their relationship to subsequent risk assessment remain insufficiently characterized. 
IBUS proved that BI-RADS descriptors provided complementary diagnostic information, with lesion margin exerting the largest influence on model performance, consistent with its established importance in breast lesion assessment~\cite{grimm2017suspicious,pfob2022new}. 
The accompanying visualizations further showed that descriptor-informed predictions were associated with attention to lesion regions corresponding to relevant imaging characteristics. 
Beyond binary benign--malignant diagnosis, accurate risk stratification is essential in breast ultrasound because it translates imaging findings into graded levels of malignancy risk that inform subsequent clinical management~\cite{zhou2026non}.
IBUS links the imaging findings with risk stratification, thereby more closely reflecting the transition from lesion characterization to BI-RADS assessment in clinical practice. 
Incorporating BI-RADS descriptors produced latent risk representations with an ordered malignancy gradient, and the resulting predicted BI-RADS grades showed benign--malignant distributions broadly consistent with radiologist-assigned categories. 
Importantly, cross-checking predicted BI-RADS grades against malignancy probabilities enabled internally discordant predictions, such as a low predicted BI-RADS category accompanied by a high malignancy probability, to be flagged for additional review. 
This does not establish that such cross-checking strategy reduces diagnostic errors in practice, but it provides a clinically interpretable means of identifying predictions for which the model's risk stratification and malignancy assessment disagree on some uncertain cases~\cite{verboom2025ai,faghani2023quantifying}. 
Thus, the value of incorporating BI-RADS knowledge may extend beyond improving prediction performance alone, as risk stratification provides intermediate outputs that are more closely aligned with established clinical assessment criteria.

The reader study further illustrates how IBUS could complement radiologists during BI-RADS assessment. 
Compared with eight radiologists, IBUS demonstrated higher sensitivity, whereas readers generally achieved higher specificity, indicating complementary diagnostic characteristics rather than a uniform advantage of either AI or human assessment. 
High specificity may help reduce unnecessary further examinations, but relatively lower sensitivity among readers would result in missed malignancies, particularly in settings where access to specialized breast imaging expertise is limited~\cite{Shete2021differences,Lawson2025Disparities}. 
When IBUS predictions were incorporated into the assessment workflow, radiologists across different experience levels showed improved overall generalization performance, with AUC increasing from 0.755 to 0.788, sensitivity from 0.679 to 0.740, and specificity from 0.830 to 0.836. 
The improvement in sensitivity without a meaningful reduction in specificity suggests that IBUS assistance primarily helped readers identify additional malignant cases while preserving their ability to recognize benign lesions.
More importantly, the reader study suggests a significant complementary role for IBUS in fine-grained risk stratification. 
Most reader-assigned BI-RADS grades remained unchanged after IBUS assistance, indicating that IBUS assistance did not broadly alter radiologists' initial assessments. 
Besides, among the assessments that were modified, the dominant change was an upgrade of malignant lesions, whereas downgrading of malignant lesions was uncommon. 
Notably, all modifications occurred in cases where the model's interpretability evidence was considered by the radiologists to focus on clinically relevant regions (see Fig.~S7 in the Appendix), suggesting that readers selectively and correctly incorporated model outputs when the accompanying evidence was clinically plausible~\cite{warren2026impact}. 
This directional pattern suggests that IBUS-derived risk information may prompt additional scrutiny of lesions that might otherwise receive a lower risk assessment, while acting primarily as a second opinion rather than replacing the reader's initial judgment~\cite{everett2026tool,yan2024domain}. 
Such support suggests that IBUS has the potential to serve as a valuable decision support tool, which may be particularly valuable in clinical environments with heterogeneous reader experience~\cite{dan2024diagnostic,abdullah2009breast}.

Several limitations should be acknowledged. 
First, despite the large-scale multicenter evaluation of IBUS in diverse breast ultrasound datasets, prospective validation is still required to further assess its performance in routine clinical workflows. 
Second, the structured BI-RADS descriptors used in this study relied on manual annotations by experienced radiologists, which are time-consuming and may limit scalability to larger clinical datasets. 
Future work will investigate the automatic extraction of BI-RADS descriptors from free-text reports using large language models, enabling a more efficient construction of such clinical knowledge. 
Finally, current framework was developed and validated specifically for breast ultrasound. 
Whether the proposed paradigm of integrating standardized radiological knowledge with visual representation learning can be generalized to other tasks, imaging modalities, or disease settings remains to be systematically investigated.

In summary, this study demonstrates that incorporating structured BI-RADS descriptors enables AI for breast ultrasound to move beyond conventional binary classification to clinically grounded decision support. 
By bridging visual representations with clinically meaningful BI-RADS descriptors, IBUS provides robust multicenter generalizability, interpretable lesion characterization, and clinically meaningful risk stratification. 
More importantly, IBUS is designed to complement radiologists by providing additional evidence for challenging diagnostic decisions rather than replacing radiologist's diagnosis and interpretation. 
Such a clinically oriented paradigm may facilitate the development of transparent and generalizable artificial intelligence systems for breast ultrasound evaluation, with high potential for translation into clinical practice.


\section{Methods}
\label{methods}

\subsection{Dataset}

In this study, we crawl a multitude of platforms for human medical US with instance annotations and open-access availability, including GitHub, Google Scholar, Kaggle, Mendeley dataset, Zenodo, Ultrasound repositories, and Data in Brief.
Meanwhile, we have also collected several clinical datasets from private institutions for external validation in real-world practice.
Patients undergoing breast imaging were retrospectively and consecutively enrolled from three medical centers to establish three independent clinical datasets: D18 from Medical Centre 1 (January 2021 to June 2021), D19 from Medical Centre 2 (January 2023 to December 2024), and D20 from Medical Centre 3 (June 2019 to June 2021).
Through this process, we arrive at a collection of 20 datasets (denoted as D1\textasciitilde D20) covering multiple clinical applications, consisting of 9,941 patients with 18,660 ultrasound images (see Table~S1 in the Appendix). 
Specifically, benign and malignant labels in all datasets were confirmed by pathology. 
Among them, D3\textasciitilde D13, D15\textasciitilde D18, and D20 additionally provide pixel-level segmentation masks, and D1, D2, D10, D17\textasciitilde D20 provide BI-RADS grades. 
Datasets D3, D5 and D16 were collected from public ultrasound repositories that provide pathology-confirmed benign or malignant labels. 
These three datasets were further curated and cleaned to remove duplicate cases, ensuring their independence compared to other datasets and consistency with the study protocol.

With the exception of D18-D20, other publicly available datasets did not provide complete BI-RADS descriptors paired with ultrasound images. 
Therefore, two radiologists were invited to review the images and finish the annotation of the BI-RADS descriptor.
Specifically, a radiologist manually annotated the corresponding BI-RADS descriptors, while ambiguous cases were further reviewed by a senior radiologist and resolved by consensus.
During the labeling process, doctors performed a blind review without benign or malignant outcomes to avoid annotation bias.
In our setting, five datasets were utilized for model training (i.e., D1 and D3\textasciitilde D6, 2,354 cases) and one for validation (D2, 846 cases).
The remaining 14 diverse datasets (i.e., D7\textasciitilde D20,  6,741 cases) were used for external validation.
All of these datasets include linear and radial probes, span multiple clinical applications and imaging techniques, allowing an exhaustive evaluation of IBUS’s generalizability.
All images were first padded along the shorter side to preserve the original aspect ratio and then resized to $224 \times 224$ pixels. 
This preprocessing avoids the geometric distortion of the lesion caused by direct resizing and better preserves the correspondence between the BI-RADS descriptors and the visual appearance.

\subsection{Model Architecture}

IBUS mainly consists of four components, including a dual-stream feature extractor, a segmentation head, a concept-informed representation coupling module and a classification head. 

\textbf{1) Dual-stream feature extractor.}
The dual-stream feature extractor is comprised of a pretrained ResNet-34~\cite{he2016deep} as visual encoder and a frozen CLIP as text encoder~\cite{radford2021learning} to extract visual and textual features from ultrasound images and BI-RADS descriptors, respectively.
Unlike conventional CLIP-based classification models, token-wise textual representations are retained to preserve fine-grained clinical semantics for subsequent cross-modal interaction. 
The resulting visual and textual features provide complementary morphological and semantic information for downstream diagnosis.
 
\textbf{2) Segmentation head.} The segmentation head is constructed to locate the lesion, adopting boundary-aware semantic contrastive learning to encourage the network to focus on clinically relevant regions. 
In this branch, an edge-enhanced convolution~\cite{cao2021shapeconv} is first introduced to strengthen boundary perception, then dense semantic contrastive learning is further utilized to improve feature discrimination under semi-supervised settings. 
Specifically, the edge-enhanced convolution receives the representations from stage1 and stage4 of ResNet-34, i.e., $\mathbf{F}=\mathrm{Conv}\big(\mathbf{K}_{B}+\mathbf{K}_{E},\mathbb{P}\big)$, where $\mathbb{P}$ represents the local feature patch.
It explicitly decomposes each convolution kernel into a base component $\mathbf{K}_{B}$ and an edge component $\mathbf{K}_{E}$, enabling the simultaneous modeling of appearance patterns and boundary structures. 
Then, a directional edge regularization is further imposed on the edge kernels,
\begin{align}
	\label{eq:edge_reg}
	\mathcal{L}_{\text{edge}}
	&=
	\sum_{d\in\mathcal{D}}
	\Big(
	\underbrace{\big|\langle \textbf{K}_E^{(d)},\mathbf{1}\rangle\big|}_{\text{(i) zero-sum}}
	+
	\underbrace{\|\textbf{K}_E^{(d)}[\mathcal{A}_d]\|_{1}}_{\text{(ii) zero axis}}
	+
	\underbrace{\|\textbf{K}_E^{(d)}+\mathcal{F}_d(\textbf{K}_E^{(d)})\|_{1}}_{\text{(iii) anti-symmetry}}
	\Big),
\end{align}
where $\langle\cdot,\cdot\rangle$ denotes the inner product of Frobenius, $\textbf{K}[\mathcal{A}_d]$ extracts the axis elements, and $\mathcal{F}_d(\cdot)$ flips the kernel along the axis $\mathcal{A}_d$ (so that paired symmetric positions are confined to be opposite). 
This regularization improves the sensitivity of boundary while suppressing noisy responses. 
The features generated by edge-enhanced convolution were then upsampled to the same resolution as the mask and concatenated to generate prediction masks via the feature fusion layer and output layer.
During training, the output masks were supervised by dice similarity coefficient loss while ground-truth masks are available.

To exploit efficient supervision from unlabeled images, we further introduce dense semantic contrastive learning to enhance foreground--background representation discrimination. 
Firstly, we defined the lesion region as foreground (i.e., $\Omega^{+}$) and the surrounding region as background (i.e., $\Omega^{+}$).
For labeled images, foreground and background regions can be extracted using ground-truth masks, whereas binarized prediction masks (threshold equal to 0.5) are used for unlabeled images. 
The pixels are then further divided into hard and easy sets according to their proximity to the boundaries of the lesion, i.e., 
$\Omega^{*}_H = \{\, \boldsymbol{\omega}_i \mid \partial(\Omega^{*},\sigma), \|i-b_i\|^2_2 \le l \,\},\Omega^{*}_E = \{\, \boldsymbol{\omega}_i \mid \partial(\Omega^{*},\sigma), \|i-b_i\|^2_2 > l \,\},*\in\{+,-\}$. $b_i$ is the closest boundary location to location $i$ from $\Omega^{*}$, $\boldsymbol{\omega}_i$ is the corresponding pixel embeddings from the feature fusion layer, $\partial(\cdot, \sigma)$ is a random sampling operator that samples a subset of a set randomly with a proportion ratio $0.2$. 
The semantic prototypes ($\mathbf{q}^{+}/\mathbf{q}^{-}$) are obtained by averaging the corresponding pixel embeddings ($\Omega^{+}$/$\Omega^{-}$).
Then, the foreground and background prototypes are contrasted with the corresponding hard/easy pixel embeddings using an InfoNCE objective~\cite{oord2018representation,wang2022contrastmask}, encouraging intra-class compactness and inter-class separation, i.e., $\mathcal{L}_{con}=
	\mathcal{L}^{\textbf{\textit{q}}^+}_{\Omega_E^{+},\Omega_E^{-}}
	+
	\mathcal{L}^{\textbf{\textit{q}}^+}_{\Omega_H^{+},\Omega_H^{-}}
	+
	\mathcal{L}^{\textbf{\textit{q}}^-}_{\Omega_E^{-},\Omega_E^{+}}
	+
	\mathcal{L}^{\textbf{\textit{q}}^-}_{\Omega_H^{-},\Omega_H^{+}}$. 
Therefore, the segmentation branch is optimized as
\begin{equation}
\mathcal{L}_{seg}
=
\mathcal{L}_{dice}
+
\lambda_c\mathcal{L}_{con}
+
\lambda_e\mathcal{L}_{edge}.
\end{equation}
where $\mathcal{L}_{dice}$ denotes the dice similarity coefficient loss.

\textbf{3) Concept-informed representation coupling module.} This module first performs fine-grained cross-modal interaction to obtain a unified visual-language representations. 
Then, based on the fused features, a GMM is introduced to model latent BI-RADS risk patterns before malignancy prediction.
Given visual features $\mathbf{F}_v$ and textual embeddings $\mathbf{F}_t$, the cross-modal interaction is formulated as:
\begin{equation}
\mathbf{F}
=
\mathrm{MHA}(\mathbf{F}_v,\mathbf{F}_t,\mathbf{F}_t)
\textcircled{c}
\mathrm{MHA}(\mathbf{F}_t,\mathbf{F}_v,\mathbf{F}_v),
\end{equation}
where $\textcircled{c}$ denotes feature concatenation. 
The fused representation is then aggregated via the average pooling operator to produce a unified embedding, i.e., $\mathbf{z}=\mathrm{AvgPool}(\mathbf{F})$, which will be clustered in GMM with six components for risk stratification. 
In GMM, the mean prototypes are iteratively updated using a memory bank and K-means clustering, i.e., $\boldsymbol{\mu}_k
=\frac{1}{|\mathcal{C}_k|}\sum_{\mathbf{z}_i\in\mathcal{C}_k}\mathbf{z}_i$
where $\mathcal{C}_k$ denotes the sets of samples in memory bank that clustered as the $k$-th component.
Then, a transformer layer with the input of mean prototypes ($\boldsymbol{\mu}_k$, as K and V in multi-head attention) and feature embedding (as Q) is introduced to generate instance-wise variance embedding, i.e., $\boldsymbol{\sigma}_{k}$.
Finally, latent representations are generated via reparameterization (i.e., $\tilde{\mathbf{z}}=\boldsymbol{\mu}_{k}+\boldsymbol{\sigma}_{k}\odot\boldsymbol{\epsilon},\boldsymbol{\epsilon}\sim\mathcal{N}(0,\mathbf{I})$).
During training, we performed multiple random samples to obtain robust output from distribution representations.

\textbf{4) Classification head.} Subsequently, all sampled features were fed into the classification head built with linear layer and optimized using cross-entropy loss:
\begin{equation}
\mathcal{L}_{cls}
=
-\sum_{m=1}^{M}\sum_{c=1}^{C}
y_c
\log
\sigma(\hat{y}_c).
\end{equation}
where $y_c$ and $\hat{y}_c$ represent the label and prediction output, respectively. 
$C$ and $M$ denote the number of classes and samples, respectively.
$\sigma$ is the Sigmoid function.
Thus, the entire framework is jointly optimized by
\begin{equation}
\mathcal{L}
=
\mathcal{L}_{cls}
+
\mathcal{L}_{seg}
=
\mathcal{L}_{cls}
+
\mathcal{L}_{dice}
+
\lambda_c\mathcal{L}_{con}
+
\lambda_e\mathcal{L}_{edge},
\end{equation}
where $\lambda_c$ and $\lambda_e$ are the coefficients of loss balance.

\subsection{Similarity-based Risk Stratification}
\label{Sim_birads_cal}
To provide clinically interpretable risk assessment, we designed a similarity-based risk stratification strategy (as shown in Fig.~\ref{risk_stra}a). 
Specifically, we first generated the reference prototype (i.e., $\mathcal{C}={\mathbf{c}_1, \mathbf{c}_2,..., \mathbf{c}_{\mathcal{D}_{r}}}$) and gaussian component (i.e., $k={1, 2,..., 6}$) of all cases from the reference data set $\mathcal{D}_{r}$ (D2 in this study). 
Given an input sample, we obtained its latent representation $\tilde{\mathbf{z}}\in\mathbb{R}^{D}$ and calculated its posterior probability to select the gaussian component with the highest probability. 
With the selected component, we performed similarity-based retrieval within the assigned gaussian component to identify clinically similar reference cases using the negative Euclidean distance:
\begin{equation}
s_i=-\left\|\tilde{\mathbf{z}}-\mathbf{c}_i\right\|_2,
\qquad
\mathbf{c}_i\in\mathcal{C}_k,
\end{equation}
where $\mathcal{C}_k$ denotes the set of reference samples belonging to the $k$-th gaussian component.
Then, similarity scores are subsequently ranked, and the Top-$N$ most similar cases are retrieved:
\begin{equation}
\mathcal{N}_{N}(X)
=
\{\mathbf{c}_{(1)},\mathbf{c}_{(2)},\ldots,\mathbf{c}_{(N)}\},
\end{equation}
where $\mathbf{c}_{(i)}$ denotes the $i$-th nearest neighbor. 
The corresponding BI-RADS categories of the retrieved cases are regarded as the candidate BI-RADS grade of the input sample.

\subsection{Reader Study}
\label{sec_reader_study}

To evaluate the clinical utility of IBUS in routine breast ultrasound diagnosis, we conducted a reader study involving eight ultrasound radiologists with different levels of experience, including four junior readers (i.e., R1\textasciitilde R4, with 1\textasciitilde 5 years experience) and four senior readers (three radiologists (i.e., R5\textasciitilde R7) are with 5\textasciitilde 10 years experience and one (i.e., R8) with >10 years experience). 
This study aimed to investigate the effectiveness of IBUS in assisting radiologists' benign--malignant diagnosis and risk stratification. 
The overall workflow is illustrated in Fig.~\ref{ai_assisted}a.

\textbf{(a) Study cohort.}
A total of 300 anonymized cases from D18 and D20 were randomly selected for evaluation. 
Each case consisted of ultrasound images together with routine clinical information, including demographic characteristics, medical history, and lesion-related clinical findings.

\textbf{(b) Independent assessment.}
Radiologists independently reviewed each case based solely on ultrasound images and clinical information, and assigned both a BI-RADS category and a benign--malignant diagnosis according to their routine clinical practice.

\textbf{(c) AI-assisted assessment.}
In this step, all cases will be reshuffled randomly.
All readers were required to provide lesion descriptors and then received explanation evidence and model-generated predictions, including CAMs, Top-$3$ risk stratification retrieval results and malignancy probability scores. 
Based on these complementary findings, radiologists could comprehensively evaluate risk stratification and benign--malignant diagnosis. 
Meanwhile, the reliability of the CAMs was independently evaluated using a three-level scoring scheme, i.e., \emph{P1} (The rate of covered lesion region is larger than 0.9), \emph{P2} (The rate of covered lesion region is larger than 0.5 but smaller than 0.9), or \emph{P3} (The rate of covered lesion region is smaller than 0.5), according to their agreement with the readers' diagnostic reasoning regions.
During the evaluation, radiologist plays a role as the primary evaluator, while IBUS as a second-stage decision-support system that provides complementary risk estimates and interpretable evidence.
All radiologists decide whether to adopt the model's predictions based on their own judgment of the evidence.
Besides, all readers were also invited to review the original BI-RADS descriptors and lesion localization maps (with same three-level scoring scheme as CAMs).
The corresponding results are shown in Fig.~S8 and S11 in the Appendix.

Throughout this study, radiologists remained blinded to the pathological diagnoses and the evaluation results of other participants, ensuring an unbiased evaluation.
Supplementary materials include a video about the IBUS-assisted process, showing how doctors use IBUS during assessment.


\subsection{Patient-level Top-$N$ BI-RADS Aggregation}
As a patient might contain multiple ultrasound images, image-level BI-RADS predictions were aggregated to obtain patient-level results. 
For Top-$1$ evaluation, the Top-$1$ BI-RADS predictions from all images of each patient were pooled, and the relative frequency of each category was calculated. 
The category with the highest frequency was assigned as the patient-level prediction. 
In the event of a two-way tie, the lower-risk BI-RADS category was selected. 
When three or more categories were tied for the highest frequency, the category corresponding to the middle risk level among the tied categories was selected.
For Top-$3$ and Top-$5$ evaluation, candidate BI-RADS categories from all images of a patient were pooled and ranked according to their frequencies. 
We then assessed whether the reference BI-RADS category was represented among the Top-$N$ candidate categories ($N=3$ or $5$). 
If the reference category was included, the case was considered to be covered at the corresponding Top-$N$ level, and the reference category was retained for category-level evaluation. 
If the reference category was not covered, a single patient-level category was assigned according to the same frequency-based aggregation and tie-breaking rules used for Top-$1$ evaluation. 
Thus, Top-$3$ and Top-$5$ represent reference-coverage-based evaluation criteria rather than independent 
single-category predictions.

\subsection{Evaluation Metrics}
We evaluated IBUS from two complementary perspectives: malignancy classification and BI-RADS risk stratification. 
For malignancy classification, the area under the receiver operating characteristic curve (AUC), accuracy (ACC), sensitivity (SEN), specificity (SPE), positive predictive value (PPV), and negative predictive value (NPV) were calculated. 
AUC was calculated from the continuous malignancy probabilities, whereas the remaining metrics were calculated after dichotomizing the predicted probabilities at a threshold of 0.5. 
To further quantify the magnitude of disagreement between predicted and reference BI-RADS categories, we defined the BI-RADS deviation (DEV) score. 
The ordered categories BI-RADS 2, 3, 4A, 4B, 4C, and 5 were sequentially encoded as 1, 2, 3, 4, 5, and 6, respectively. For the $i$th case, the DEV score was defined as:
\begin{equation}
    \mathrm{DEV}_{i}
    =
    \left| r_{i} - \hat{r}_{i} \right|,
\end{equation}
where $r_i$ and $\hat{r}_i$ denote the encoded reference and evaluated BI-RADS 
categories, respectively. 
A DEV score of 0 indicates exact agreement, whereas larger values indicate progressively greater ordinal disagreement.
The mean DEV score was calculated as:
\begin{equation}
    \overline{\mathrm{DEV}}
    =
    \frac{1}{N}
    \sum_{i=1}^{N}
    \left| r_{i} - \hat{r}_{i} \right|,
\end{equation}
where $N$ denotes the number of evaluated cases. 
Mean DEV scores were calculated for the overall cohort and separately for each reference BI-RADS category (2, 3, 4A, 4B, 4C, and 5) under the Top-$1$, Top-$3$, and Top-$5$ evaluation criteria. 
A lower mean DEV indicates closer ordinal agreement with the reference BI-RADS grades; values below 1 correspond to an average discrepancy of less than one BI-RADS category step.

\subsection{Implementation Details}

The CLIP text encoder remained frozen throughout training. 
Since the length of BI-RADS descriptors typically contained fewer than 35 tokens, considerably shorter than CLIP's maximum sequence length of 77, trailing padding tokens were dynamically removed within each mini-batch to reduce unnecessary computation.
For GMM, a first-in-first-out (FIFO) memory bank with a capacity of 512 samples was maintained to store samples embeddings. 
Gaussian prototypes were initialized randomly and updated using K-means clustering on the memory bank embeddings. 
During the first five epochs, the prototypes were updated after every epoch using exponential moving average (EMA); thereafter, updates were performed every five epochs to improve training stability. 
The model was trained using the Adam optimizer with an initial learning rate of $1\times10^{-4}$, which was decayed by a factor of 0.8 every 10 epochs. 
Training was performed for 150 epochs with a batch size of 28. 
The loss weights were empirically set to $\lambda_e=10$ and $\lambda_c=0.1$. 
The model with the best validation performance was selected for all subsequent experiments.
All compared methods were reproduced according to their original publications and publicly available implementations. 
For segmentation comparison, all competing models were trained in a fully supervised manner using datasets with pixel-level annotations, and the optimal models were selected using a validation set randomly sampled from 10\% of the training data.
All metrics are calculated based on patient-level prediction. 
For patients with multiple images, the probability of malignancy is based on the average across all images.
All models were implemented in PyTorch 2.4.1 and CUDA 12.1 based on Python 3.8.19 and trained on a workstation equipped with a NVIDIA RTX 4090 GPU (24 GB).
Scikit-learn (v1.3.2) was used for evaluation.

\subsection{Code availability}
\label{code_aval}
We provided the IBUS-assisted demo video at \textcolor{blue}{https://github.com/zsz1997/IBUS}, where source code will be also released. 

\subsection{Acknowledgements}
This work was supported in part by the National Natural Science Foundation of China under Grant 82402363, Grant U22A20350 and Grant 82472056, in part by the Basic and Applied Basic Research Foundation of Guangdong Province under Grant 2025A1515011205 and Grant 2024A1515012004, in part by the Guangdong Provincial Key Laboratory of Medical Image Processing under Grant No. 2020B1212060039, in part by the Science and Technology Program of Guangzhou under Grant No. 2025A04J4019, in part by the Guangdong Association for Science and Technology Young Talents Training Program under Grant SKXRC2025372, in part by the Special Funds for the Cultivation of Guangdong College Students’ Scientific and Technological Innovation under Grant No. pdjh2023a0099 and Grant No. pdjh2024a087, and in part by the Research Support Fund for High-level Talents of Southern Medical University under Grant No. G624300273.

\subsection{Supplementary information}
Supplementary information, including Supplementary Tables and Supplementary Figures, is provided with this preprint.

\bibliographystyle{naturemag}
\bibliography{references}

@article{heer2020global,
  title={Global burden and trends in premenopausal and postmenopausal breast cancer: a population-based study},
  author={Heer, Emily and Harper, Andrew and Escandor, Noah and Sung, Hyuna and McCormack, Valerie and Fidler-Benaoudia, Miranda M},
  journal={The Lancet Global Health},
  volume={8},
  number={8},
  pages={e1027--e1037},
  year={2020},
  publisher={Elsevier}
}

@article{sung2026global,
  title={Global cancer statistics 2024: GLOBOCAN estimates of incidence and mortality worldwide for 34 cancers in 186 countries},
  author={Sung, Hyuna and Filho, Adalberto M and Laversanne, Mathieu and Ferlay, Jacques and Siegel, Rebecca L and Soerjomataram, Isabelle and Jemal, Ahmedin and Bray, Freddie},
  journal={CA: A Cancer Journal for Clinicians},
  volume={76},
  number={4},
  pages={e70090},
  year={2026},
  publisher={Wiley Online Library}
}

@article{xiong2025breast,
  title={Breast cancer: pathogenesis and treatments},
  author={Xiong, Xin and Zheng, Le-Wei and Ding, Yu and Chen, Yu-Fei and Cai, Yu-Wen and Wang, Lei-Ping and Huang, Liang and Liu, Cui-Cui and Shao, Zhi-Ming and Yu, Ke-Da},
  journal={Signal Transduction and Targeted Therapy},
  volume={10},
  number={1},
  pages={49},
  year={2025},
  publisher={Nature Publishing Group UK London}
}

@article{harada2021evaluation,
	title={Evaluation of adjunctive ultrasonography for breast cancer detection among women aged 40-49 years with varying breast density undergoing screening mammography: a secondary analysis of a randomized clinical trial},
	author={Harada-Shoji, Narumi and Suzuki, Akihiko and Ishida, Takanori and Zheng, Ying-Fang and Narikawa-Shiono, Yoko and Sato-Tadano, Akiko and Ohta, Rie and Ohuchi, Noriaki},
	journal={JAMA Network Open},
	volume={4},
	number={8},
	pages={e2121505--e2121505},
	year={2021},
	publisher={American Medical Association}
}

@article{gilbert2025comparison,
	title={Comparison of supplemental breast cancer imaging techniques—interim results from the BRAID randomised controlled trial},
	author={Gilbert, Fiona J and Payne, Nicholas R and Allajbeu, Iris and Yit, Lim and Vinnicombe, Sarah and Lyburn, Iain and Sharma, Nisha and Teh, Will and James, Jonathan and Seth, Archana and others},
	journal={The Lancet},
	volume={405},
	number={10493},
	pages={1935--1944},
	year={2025},
	publisher={Elsevier}
}

@article{qian2021prospective,
	title={Prospective assessment of breast cancer risk from multimodal multiview ultrasound images via clinically applicable deep learning},
	author={Qian, Xuejun and Pei, Jing and Zheng, Hui and Xie, Xinxin and Yan, Lin and Zhang, Hao and Han, Chunguang and Gao, Xiang and Zhang, Hanqi and Zheng, Weiwei and others},
	journal={Nature Biomedical Engineering},
	volume={5},
	number={6},
	pages={522--532},
	year={2021},
	publisher={Nature Publishing Group UK London}
}

@article{luo2024deep,
	title={Deep learning in breast cancer imaging: A decade of progress and future directions},
	author={Luo, Luyang and Wang, Xi and Lin, Yi and Ma, Xiaoqi and Tan, Andong and Chan, Ronald and Vardhanabhuti, Varut and Chu, Winnie CW and Cheng, Kwang-Ting and Chen, Hao},
	journal={IEEE Reviews in Biomedical Engineering},
	year={2024},
	publisher={IEEE}
}

@article{dan2024diagnostic,
	title={Diagnostic performance of deep learning in ultrasound diagnosis of breast cancer: a systematic review},
	author={Dan, Qing and Xu, Ziting and Burrows, Hannah and Bissram, Jennifer and Stringer, Jeffrey SA and Li, Yingjia},
	journal={NPJ Precision Oncology},
	volume={8},
	number={1},
	pages={21},
	year={2024},
	publisher={Nature Publishing Group UK London}
}

@article{chang2025hallmarks,
	title={Hallmarks of artificial intelligence contributions to precision oncology},
	author={Chang, Tian-Gen and Park, Seongyong and Sch{\"a}ffer, Alejandro A and Jiang, Peng and Ruppin, Eytan},
	journal={Nature Cancer},
	pages={1--15},
	year={2025},
	publisher={Nature Publishing Group US New York}
}

@article{atlas2013breast,
	title={Breast imaging reporting and data system},
	author={Atlas, ACR BI-RADS},
	journal={American College of Radiology},
	year={2013}
}

@article{shen2021artificial,
	title={Artificial intelligence system reduces false-positive findings in the interpretation of breast ultrasound exams},
	author={Shen, Yiqiu and Shamout, Farah E and Oliver, Jamie R and Witowski, Jan and Kannan, Kawshik and Park, Jungkyu and Wu, Nan and Huddleston, Connor and Wolfson, Stacey and Millet, Alexandra and others},
	journal={Nature Communications},
	volume={12},
	number={1},
	pages={5645},
	year={2021},
	publisher={Nature Publishing Group UK London}
}

@article{myers2015benefits,
	title={Benefits and harms of breast cancer screening: a systematic review},
	author={Myers, Evan R and Moorman, Patricia and Gierisch, Jennifer M and Havrilesky, Laura J and Grimm, Lars J and Ghate, Sujata and Davidson, Brittany and Mongtomery, Ranee Chatterjee and Crowley, Matthew J and McCrory, Douglas C and others},
	journal={JAMA},
	volume={314},
	number={15},
	pages={1615--1634},
	year={2015},
	publisher={American Medical Association}
}

@article{qian2025multimodal,
	title={A multimodal machine learning model for the stratification of breast cancer risk},
	author={Qian, Xuejun and Pei, Jing and Han, Chunguang and Liang, Zhiying and Zhang, Gaosong and Chen, Na and Zheng, Weiwei and Meng, Fanlun and Yu, Dongsheng and Chen, Yixuan and others},
	journal={Nature Biomedical Engineering},
	volume={9},
	number={3},
	pages={356--370},
	year={2025},
	publisher={Nature Publishing Group UK London}
}

@article{lecun2015deep,
	title={Deep learning},
	author={LeCun, Yann and Bengio, Yoshua and Hinton, Geoffrey},
	journal={Nature},
	volume={521},
	number={7553},
	pages={436--444},
	year={2015},
	publisher={Nature Publishing Group UK London}
}

@article{bi2019artificial,
  title={Artificial intelligence in cancer imaging: clinical challenges and applications},
  author={Bi, Wenya Linda and Hosny, Ahmed and Schabath, Matthew B and Giger, Maryellen L and Birkbak, Nicolai J and Mehrtash, Alireza and Allison, Tavis and Arnaout, Omar and Abbosh, Christopher and Dunn, Ian F and others},
  journal={CA: A Cancer Journal for Clinicians},
  volume={69},
  number={2},
  pages={127--157},
  year={2019},
  publisher={Wiley Online Library}
}

@article{mckinney2020international,
  title={International evaluation of an AI system for breast cancer screening},
  author={McKinney, Scott Mayer and Sieniek, Marcin and Godbole, Varun and Godwin, Jonathan and Antropova, Natasha and Ashrafian, Hutan and Back, Trevor and Chesus, Mary and Corrado, Greg S and Darzi, Ara and others},
  journal={Nature},
  volume={577},
  number={7788},
  pages={89--94},
  year={2020},
}

@article{jiang2017artificial,
	title={Artificial intelligence in healthcare: past, present and future},
	author={Jiang, Fei and Jiang, Yong and Zhi, Hui and Dong, Yi and Li, Hao and Ma, Sufeng and Wang, Yilong and Dong, Qiang and Shen, Haipeng and Wang, Yongjun},
	journal={Stroke and Vascular Neurology},
	volume={2},
	number={4},
	year={2017},
	publisher={BMJ Specialist Journals}
}

@article{leibig2022combining,
  title={Combining the strengths of radiologists and AI for breast cancer screening: a retrospective analysis},
  author={Leibig, Christian and Brehmer, Moritz and Bunk, Stefan and Byng, Danalyn and Pinker, Katja and Umutlu, Lale},
  journal={The Lancet Digital Health},
  volume={4},
  number={7},
  pages={e507--e519},
  year={2022},
  publisher={Elsevier}
}

@article{zhou2023transformer,
	title={A transformer-based representation-learning model with unified processing of multimodal input for clinical diagnostics},
	author={Zhou, Hong-Yu and Yu, Yizhou and Wang, Chengdi and Zhang, Shu and Gao, Yuanxu and Pan, Jia and Shao, Jun and Lu, Guangming and Zhang, Kang and Li, Weimin},
	journal={Nature Biomedical Engineering},
	volume={7},
	number={6},
	pages={743--755},
	year={2023},
	publisher={Nature Publishing Group UK London}
}

@article{yao2023virtual,
	title={Virtual elastography ultrasound via generative adversarial network for breast cancer diagnosis},
	author={Yao, Zhao and Luo, Ting and Dong, Yijie and Jia, Xiaohong and Deng, Yinhui and Wu, Guoqing and Zhu, Ying and Zhang, Jingwen and Liu, Juan and Yang, Lichun and others},
	journal={Nature Communications},
	volume={14},
	number={1},
	pages={788},
	year={2023},
	publisher={Nature Publishing Group UK London}
}

@article{yu2026foundation,
  title={A foundation generative model for breast ultrasound image analysis},
  author={Yu, Haojun and Li, Youcheng and Zhang, Nan and Niu, Zihan and Gong, Xuantong and Luo, Yanwen and Ye, Haotian and He, Siyu and Wu, Quanlin and Qin, Wangyan and others},
  journal={Nature Biomedical Engineering},
  pages={1--13},
  year={2026},
  publisher={Nature Publishing Group UK London}
}

@article{jiao2024usfm,
	title={Usfm: A universal ultrasound foundation model generalized to tasks and organs towards label efficient image analysis},
	author={Jiao, Jing and Zhou, Jin and Li, Xiaokang and Xia, Menghua and Huang, Yi and Huang, Lihong and Wang, Na and Zhang, Xiaofan and Zhou, Shichong and Wang, Yuanyuan and others},
	journal={Medical Image Analysis},
	volume={96},
	pages={103202},
	year={2024},
	publisher={Elsevier}
}

@article{jiang2025pretraining,
	title={From pretraining to privacy: federated ultrasound foundation model with self-supervised learning},
	author={Jiang, Yuncheng and Feng, Chun-Mei and Ren, Jinke and Wei, Jun and Zhang, Zixun and Hu, Yiwen and Liu, Yunbi and Sun, Rui and Tang, Xuemei and Du, Juan and others},
	journal={NPJ Digital Medicine},
	volume={8},
	number={1},
	pages={714},
	year={2025},
	publisher={Nature Publishing Group UK London}
}

@article{meyer2025ultrasam,
	title={Ultrasam: a foundation model for ultrasound using large open-access segmentation datasets},
	author={Meyer, Adrien and Murali, Aditya and Zarin, Farahdiba and Mutter, Didier and Padoy, Nicolas},
	journal={International Journal of Computer Assisted Radiology and Surgery},
	pages={1--10},
	year={2025},
	publisher={Springer}
}

@inproceedings{he2016deep,
	title={Deep residual learning for image recognition},
	author={He, Kaiming and Zhang, Xiangyu and Ren, Shaoqing and Sun, Jian},
	booktitle={Proceedings of the IEEE/CVF Conference on Computer Vision and Pattern Recognition (CVPR)},
	pages={770--778},
	year={2016}
}

@inproceedings{radford2021learning,
	title={Learning transferable visual models from natural language supervision},
	author={Radford, Alec and Kim, Jong Wook and Hallacy, Chris and Ramesh, Aditya and Goh, Gabriel and Agarwal, Sandhini and Sastry, Girish and Askell, Amanda and Mishkin, Pamela and Clark, Jack and others},
	booktitle={Proceedings of the International Conference on Machine Learning (ICML)},
	pages={8748--8763},
	year={2021},
	organization={PmLR}
}

@article{yi2024idc,
	title={IDC-Net: breast cancer classification network based on BI-RADS 4},
	author={Yi, Sanli and Chen, Ziyan and She, Furong and Wang, Tianwei and Yang, Xuelian and Chen, Dong and Luo, Xiaomao},
	journal={Pattern Recognition},
	volume={150},
	pages={110323},
	year={2024},
	publisher={Elsevier}
}

@article{mo2023hover,
	title={Hover-trans: Anatomy-aware hover-transformer for roi-free breast cancer diagnosis in ultrasound images},
	author={Mo, Yuhao and Han, Chu and Liu, Yu and Liu, Min and Shi, Zhenwei and Lin, Jiatai and Zhao, Bingchao and Huang, Chunwang and Qiu, Bingjiang and Cui, Yanfen and others},
	journal={IEEE Transactions on Medical Imaging},
	volume={42},
	number={6},
	pages={1696--1706},
	year={2023},
	publisher={IEEE}
}

@article{aumente2025multi,
	title={A multi-task framework for breast cancer segmentation and classification in ultrasound imaging},
	author={Aumente-Maestro, Carlos and D{\'\i}ez, Jorge and Remeseiro, Beatriz},
	journal={Computer Methods and Programs in Biomedicine},
	volume={260},
	pages={108540},
	year={2025},
	publisher={Elsevier}
}

@article{vega2025overcoming,
  title={Overcoming barriers in the use of artificial intelligence in point of care ultrasound},
  author={Vega, Roberto and Dehghan, Masood and Nagdev, Arun and Buchanan, Brian and Kapur, Jeevesh and Jaremko, Jacob L and Zonoobi, Dornoosh},
  journal={NPJ Digital Medicine},
  volume={8},
  number={1},
  pages={213},
  year={2025},
  publisher={Nature Publishing Group UK London}
}

@article{zhao2026clinic,
  title={Clinic-aligned dual distillation of video and image foundation models for automated breast cancer US diagnosis},
  author={Zhao, Chengqian and Dong, Yijie and Xie, Xuan and Wu, Guoqing and Yin, Feiyu and Zeng, Wenwen and Song, Pengfei and Wu, Yonghuang and Fu, Lina and Nie, Bingqian and others},
  journal={Radiology: Artificial Intelligence},
  volume={8},
  number={3},
  pages={e250600},
  year={2026},
  publisher={Radiological Society of North America}
}

@article{Pfob2022,
  title={The importance of multi-modal imaging and clinical information for humans and AI-based algorithms to classify breast masses (INSPiRED 003): an international, multicenter analysis},
  author={Pfob, Andr{\'e} and Sidey-Gibbons, Chris and Barr, Richard G and Duda, Volker and Alwafai, Zaher and Balleyguier, Corinne and Clevert, Dirk-Andr{\'e} and Fastner, Sarah and Gomez, Christina and Goncalo, Manuela and others},
  journal={European Radiology},
  volume={32},
  number={6},
  pages={4101--4115},
  year={2022},
  publisher={Springer}
}

@article{Luo2025,
  title={A large model for non-invasive and personalized management of breast cancer from multiparametric MRI},
  author={Luo, Luyang and Wu, Mingxiang and Li, Mei and Xin, Yi and Wang, Qiong and Vardhanabhuti, Varut and Chu, Winnie CW and Li, Zhenhui and Zhou, Juan and Rajpurkar, Pranav and others},
  journal={Nature Communications},
  volume={16},
  number={1},
  pages={3647},
  year={2025},
  publisher={Nature Publishing Group UK London}
}

@article{Ali2025,
  title={Exploring AI approaches for breast cancer detection and diagnosis: A review Article},
  author={Ali, Akbar and Alghamdi, Mansoor and Marzuki, Shahira Sofea and Tengku Din, Tengku Ahmad Damitri Al Astani and Yamin, Muhamad Syahmi and Alrashidi, Malek and Alkhazi, Ibrahim S and Ahmed, Naveed},
  journal={Breast Cancer: Targets and Therapy},
  pages={927--947},
  year={2025},
  publisher={Taylor \& Francis}
}

@article{li2024ultrasound,
  title={Ultrasound report generation with cross-modality feature alignment via unsupervised guidance},
  author={Li, Jun and Su, Tongkun and Zhao, Baoliang and Lv, Faqin and Wang, Qiong and Navab, Nassir and Hu, Ying and Jiang, Zhongliang},
  journal={IEEE Transactions on Medical Imaging},
  volume={44},
  number={1},
  pages={19--30},
  year={2024},
  publisher={IEEE}
}

@article{alsharid2025public,
  title={On the public dissemination and open sourcing of ultrasound resources, datasets and deep learning models},
  author={Alsharid, Mohammad and Guo, Xiaoqing and Men, Qianhui and Saha, Pramit and Mishra, Divyanshu and Ahuja, Rahul and Ouyang, Cheng and Noble, J Alison},
  journal={NPJ Digital Medicine},
  volume={8},
  number={1},
  pages={777},
  year={2025},
  publisher={Nature Publishing Group UK London}
}

@inproceedings{le2026u2,
  title={U2-bench: Benchmarking large vision-language models on ultrasound understanding},
  author={Le, Anjie and Liu, Henan and Wang, Yue and Liu, Zhenyu and Zhu, Rongkun and Weng, Taohan and Yu, Jinze and Wang, Boyang and Wu, Yalun and Yan, Kaiwen and others},
  booktitle={Proceedings of the International Conference on Learning Representations (ICLR)},
  volume={2026},
  pages={8174--8225},
  year={2026}
}

@article{grimm2017suspicious,
	title={Suspicious breast calcifications undergoing stereotactic biopsy in women ages 70 and over: breast cancer incidence by BI-RADS descriptors},
	author={Grimm, Lars J and Johnson, David Y and Johnson, Karen S and Baker, Jay A and Soo, Mary Scott and Hwang, E Shelley and Ghate, Sujata V},
	journal={European Radiology},
	volume={27},
	number={6},
	pages={2275--2281},
	year={2017},
	publisher={Springer}
}

@article{pfob2022new,
	title={A new practical decision rule to better differentiate BI-RADS 3 or 4 breast masses on breast ultrasound},
	author={Pfob, Andr{\'e} and Barr, Richard G and Duda, Volker and B{\"u}sch, Christopher and Bruckner, Thomas and Spratte, Julia and Nees, Juliane and Togawa, Riku and Ho, Chi and Fastner, Sarah and others},
	journal={Journal of Ultrasound in Medicine},
	volume={41},
	number={2},
	pages={427--436},
	year={2022},
	publisher={Wiley Online Library}
}

@article{Shete2021differences,
    author = {Shete, Sanjay and Deng, Yangyang and Shannon, Jackilen and Faseru, Babalola and Middleton, Deirdre and Iachan, Ronaldo and Bernardo, Brittany and Balkrishnan, Rajesh and Kim, Sunny Jung and Huang, Bin and Millar, Morgan M. and Fuemmler, Bernard and Jensen, Jakob D. and Mendoza, Jason A. and Hu, Jinxiang and Lazovich, DeAnn and Robertson, Linda and Demark-Wahnefried, Wendy and Paskett, Electra D. and Rural Workgroup of the Population Health Assessment in Cancer Center Catchment Areas Initiative},
    title = {Differences in Breast and Colorectal Cancer Screening Adherence Among Women Residing in Urban and Rural Communities in the United States},
    journal = {JAMA Network Open},
    volume = {4},
    number = {10},
    pages = {e2128000-e2128000},
    year = {2021}
}

@article{Lawson2025Disparities,
author = {Lawson, Marissa B. and Zhu, Weiwei and Miglioretti, Diana L. and Onega, Tracy and Henderson, Louise M. and Rauscher, Garth H. and Kerlikowske, Karla and Sprague, Brian L. and Bowles, Erin J. A. and O’Meara, Ellen S. and Tosteson, Anna N. A. and diFlorio-Alexander, Roberta M. and Hubbard, Rebecca A. and Lee, Janie M. and Lee, Christoph I.},
title = {Disparities in Standard-of-Care, Advanced, and Same-Day Diagnostic Services among Patients with Abnormal Screening Mammography},
journal = {Radiology},
volume = {314},
number = {2},
pages = {e241673},
year = {2025},
doi = {10.1148/radiol.241673},
}

@article{abdullah2009breast,
  title={Breast imaging reporting and data system lexicon for US: interobserver agreement for assessment of breast masses},
  author={Abdullah, Nouf and Mesurolle, Beno{\^\i}t and El-Khoury, Mona and Kao, Ellen},
  journal={Radiology},
  volume={252},
  number={3},
  pages={665--672},
  year={2009}
}

@inproceedings{cao2021shapeconv,
	title={Shapeconv: Shape-aware convolutional layer for indoor rgb-d semantic segmentation},
	author={Cao, Jinming and Leng, Hanchao and Lischinski, Dani and Cohen-Or, Daniel and Tu, Changhe and Li, Yangyan},
	booktitle={Proceedings of the IEEE/CVF International Conference on Computer Vision (ICCV)},
	pages={7088--7097},
	year={2021}
}

@inproceedings{wang2022contrastmask,
  title={Contrastmask: Contrastive learning to segment every thing},
  author={Wang, Xuehui and Zhao, Kai and Zhang, Ruixin and Ding, Shouhong and Wang, Yan and Shen, Wei},
  booktitle={Proceedings of the IEEE/CVF Conference on Computer Vision and Pattern Recognition (CVPR)},
  pages={11594--11603},
  year={2022},
  organization={IEEE}
}

@article{oord2018representation,
  title={Representation learning with contrastive predictive coding},
  author={Oord, Aaron van den and Li, Yazhe and Vinyals, Oriol},
  journal={arXiv preprint arXiv:1807.03748},
  year={2018}
}

@article{everett2026tool,
  title={From tool to teammate in a randomized controlled trial of clinician-AI collaborative workflows for diagnosis},
  author={Everett, Selin S and Bunning, Bryan J and Jain, Priyank and Lopez, Ivan and Agarwal, Anup and Desai, Manisha and Gallo, Robert and Goh, Ethan and Kadiyala, Vinay B and Kanjee, Zahir and others},
  journal={NPJ Digital Medicine},
  volume={9},
  number={1},
  pages={409},
  year={2026},
  publisher={Nature Publishing Group UK London}
}

@article{zhou2026non,
  title={A non-invasive end-to-end intelligent assistance system for breast ultrasound},
  author={Zhou, Jun and Si, Pilei and Zhang, Yan and Song, Jianqiong and He, Tao and Liu, Qin and Chen, Shaoqi and Tu, Mengjie and Wang, Han and Guo, Qiang and others},
  journal={Nature Communications},
  year={2026},
  publisher={Nature Publishing Group UK London}
}

@article{goldhirsch2013personalizing,
  title={Personalizing the treatment of women with early breast cancer: highlights of the St Gallen International Expert Consensus on the Primary Therapy of Early Breast Cancer 2013},
  author={Goldhirsch, Aron and Winer, Eric P and Coates, Alan S and Gelber, Richard D and Piccart-Gebhart, Martine and Th{\"u}rlimann, Beat and Senn, H-J and Albain, Kathy S and Andr{\'e}, Fabrice and Bergh, Jonas and others},
  journal={Annals of Oncology},
  volume={24},
  number={9},
  pages={2206--2223},
  year={2013},
  publisher={Elsevier}
}

@article{verboom2025ai,
  title={AI Should Read Mammograms Only When Confident: A Hybrid Breast Cancer Screening Reading Strategy},
  author={Verboom, Sarah D and Kroes, Jaap and Pires, Santiago and Broeders, Mireille JM and Sechopoulos, Ioannis},
  journal={Radiology},
  volume={316},
  number={2},
  pages={e242594},
  year={2025},
  publisher={Radiological Society of North America}
}

@article{faghani2023quantifying,
  title={Quantifying uncertainty in deep learning of radiologic images},
  author={Faghani, Shahriar and Moassefi, Mana and Rouzrokh, Pouria and Khosravi, Bardia and Baffour, Francis I and Ringler, Michael D and Erickson, Bradley J},
  journal={Radiology},
  volume={308},
  number={2},
  pages={e222217},
  year={2023},
  publisher={Radiological Society of North America}
}

@article{yan2024domain,
  title={A domain knowledge-based interpretable deep learning system for improving clinical breast ultrasound diagnosis},
  author={Yan, Lin and Liang, Zhiying and Zhang, Hao and Zhang, Gaosong and Zheng, Weiwei and Han, Chunguang and Yu, Dongsheng and Zhang, Hanqi and Xie, Xinxin and Liu, Chang and others},
  journal={Communications Medicine},
  volume={4},
  number={1},
  pages={90},
  year={2024},
  publisher={Nature Publishing Group UK London}
}

@article{warren2026impact,
  title={Impact of using artificial intelligence as a second reader in breast screening including arbitration},
  author={Warren, Lucy M and Venton, Jenny and Young, Kenneth C and Halling-Brown, Mark and Kelly, Christopher J and Wilson, Marc and Morigami, Megumi and Khoo, Lisanne and Cunningham, Deborah and Sidebottom, Richard and others},
  journal={Nature Cancer},
  volume={7},
  number={3},
  pages={507--521},
  year={2026},
  publisher={Nature Publishing Group US New York}
}

\setcounter{figure}{0}
\renewcommand{\thefigure}{S\arabic{figure}}

\setcounter{table}{0}
\renewcommand{\thetable}{S\arabic{table}}

\newpage

\center{\textbf{\large{SUPPLEMENTARY APPENDIX}}
	\\ to \\A Concept-informed Interpretable Artificial Intelligence Framework for Breast Ultrasound Diagnosis\\ by Shengzhou Zhong et al}

\begin{table*}[h]
	\footnotesize
	\centering
	\caption{\textbf{Summary of study data used in this work.}
		The table reports the number of images, all patients, benign patients and malignant patients in each dataset.}
	\label{dataset}
	\begin{tabularx}{\columnwidth}{c  Y Y Y Y}
		\hline
		\textbf{Dataset}  & \textbf{Images} & \textbf{Patients} & \textbf{Benign Patients} & \textbf{Malignant Patients}  \\
		\hline
		D1  & 1,559 & 1,559 & 443 & 1,116  \\
		D2  & 846   & 846   & 443 & 403   \\
		D3  & 2,023 & 364   & 171 & 193 \\
		D4  & 200   & 100   & 48  & 52 \\
		D5  & 168   & 168   & 68  & 100  \\
		D6  & 163   & 163   & 109 & 54   \\
		D7  & 201   & 201   & 178 & 23    \\
		D8  & 232   & 232   & 109 & 123   \\
		D9  & 250   & 50    & 20  & 30     \\
		D10 & 252   & 252   & 154 & 98 \\
		D11 & 261   & 261   & 190 & 71     \\
		D12 & 260   & 260   & 170 & 90\\
		D13 & 647   & 647   & 437 & 210    \\
		D14 & 781   & 248   & 145 & 103   \\
		D15 & 927   & 816   & 560 & 256  \\
		D16 & 2,364 & 544   & 246 & 298   \\
		D17 & 1,872 & 1,061 & 722 & 339   \\
		D18 & 404   & 166   & 80  & 86     \\
		D19 & 2436  & 816   & 411  & 405  \\
		D20 & 2,814 & 1,187 & 593 & 594    \\
		\hline
		\textbf{Total} & \textbf{18,660} & \textbf{9,941} & \textbf{5,297} & \textbf{4,644}  \\
		\hline
	\end{tabularx}
	\vspace{-2em}

\end{table*}

\begin{table}[t]
	\centering
	\caption{\textbf{BI-RADS descriptors and corresponding abbreviations.}}
	\label{birads_descriptor}
	\begin{tabularx}{\columnwidth}{l X l}
		\hline
		\textbf{Category} & \multicolumn{2}{c}{\textbf{Descriptor} (\textbf{Abbreviation})} \\
		\hline
		Orientation
		& Parallel (OR1) & Unparallel (OR2) \\
		\hline
		\multirow{2}{*}{Shape} 
		& Round (SH1) & Oval (SH2) \\
		& Irregular (SH3) & \\
		\hline
		\multirow{2}{*}{Calcifications} 
		& No calcifications (CA1) & Calcifications in mass (CA2) \\
		& Calcifications outside mass (CA3) & Intraductal calcifications (CA4) \\
		\hline
		\multirow{2}{*}{Posterior features} 
		& No posterior features (PF1) & Enhanced (PF2) \\
		& Shadowing (PF3) & Combined (PF4) \\
		\hline
		\multirow{3}{*}{Echo pattern} 
		& Anechoic (EC1) & Hypoechoic (EC2) \\
		& Isoechoic (EC3) & Hyperechoic (EC4) \\
		& Complex cystic or solid (EC5) & Heterogeneous (EC6) \\
		\hline
		\multirow{4}{*}{Margin} 
		& Circumscribed (ED1) & Indistinct (ED2) \\
		& Microlobulated (ED3) & Angular (ED4) \\
		& Spiculated (ED5) & Architectural distortion (ED6) \\
		& Duct dilation and intraductal mass (ED7) &\\
		\hline
	\end{tabularx}
\end{table}

\begin{table}[t]
	\centering
	\caption{Diagnostic performance of IBUS, junior readers, senior readers without and with IBUS assistance.}
	\label{tab:reader_IBUS}
	\scriptsize   
	\begin{tabularx}{\columnwidth}{c YYYYYYYY}
		\hline
		\textbf{Method} & \textbf{AUC} & \textbf{ACC} & \textbf{SEN} & \textbf{SPE} & \textbf{PPV} & \textbf{NPV} \\
		\hline
		IBUS
		& \makecell{0.849 \\(0.802-0.892)}	
		& \makecell{0.780 \\(0.733-0.823)}	
		& \makecell{0.846 \\(0.790-0.902)}	
		& \makecell{0.708 \\(0.632-0.777)}	
		& \makecell{0.759 \\(0.694-0.818)}	
		& \makecell{0.810 \\(0.740-0.876)}\\
		
		Readers &&&&&&\\
		
		ALL
		& \makecell{0.755* \\(0.724-0.785)}	
		& \makecell{0.752* \\(0.718-0.784)}	
		& \makecell{0.679* \\(0.628-0.727)}	
		& \makecell{0.830* \\(0.791-0.867)}	
		& \makecell{0.817* \\(0.769-0.862)}	
		& \makecell{0.709* \\(0.649-0.764)} \\
		
		Junior
		& \makecell{0.722* \\(0.687-0.757)}	
		& \makecell{0.718* \\(0.680-0.756)}	
		& \makecell{0.623* \\(0.565-0.678)}	
		& \makecell{0.821* \\(0.776-0.865)}	
		& \makecell{0.795* \\(0.740-0.850)}	
		& \makecell{0.670* \\(0.608-0.729)} \\
		
		Senior
		& \makecell{0.787* \\(0.753-0.820)}	
		& \makecell{0.785 \\(0.750-0.819)}	
		& \makecell{0.736* \\(0.680-0.787)}	
		& \makecell{0.839* \\(0.797-0.878)}	
		& \makecell{0.838* \\(0.792-0.882)}	
		& \makecell{0.748* \\(0.687-0.805)} \\
		
		Readers with IBUS assistance &&&&&&\\
		
		ALL
		& \makecell{0.788† \\(0.755-0.818)}	
		& \makecell{0.786† \\(0.752-0.817)}	
		& \makecell{0.740† \\(0.690-0.787)}	
		& \makecell{0.836 \\(0.794-0.875)}	
		& \makecell{0.832† \\(0.785-0.877)}	
		& \makecell{0.750† \\(0.692-0.803)} \\
		
		Junior
		& \makecell{0.756† \\(0.721-0.791)}	
		& \makecell{0.754† \\(0.718-0.789)}	
		& \makecell{0.702† \\(0.647-0.757)}	
		& \makecell{0.811 \\(0.761-0.856)}	
		& \makecell{0.801 \\(0.749-0.857)}	
		& \makecell{0.715† \\(0.653-0.773)} \\

		Senior
		& \makecell{0.819† \\(0.786-0.851)}	
		& \makecell{0.818† \\(0.782-0.851)}	
		& \makecell{0.777† \\(0.724-0.828)}	
		& \makecell{0.861 \\(0.816-0.902)}	
		& \makecell{0.862 \\(0.816-0.906)}	
		& \makecell{0.784† \\(0.726-0.837)}
		\\
		
		\hline
	\end{tabularx}
	\vspace{0.6em}
	\parbox{\linewidth}{
		\raggedright
		\footnotesize
		\setlength{\baselineskip}{8pt}
		Numbers within parentheses indicate 95\% Confidence Intervals. * Significant differences between readers and IBUS ($P< 0.05$) were calculated using two-sided DeLong test or two-sided Wilcoxon signed-rank test without adjustment. † Significant differences between readers without and with IBUS assistance ($P< 0.05$) were calculated using two-sided DeLong test or two-sided Wilcoxon signed-rank test without adjustment.}
\end{table}

\begin{table}[t]
	\centering
	\caption{Diagnostic performance of individual readers without and with IBUS assistance.}
	\label{tab:READER_w/oIBUS_individual}
	\scriptsize   
	\begin{tabularx}{\columnwidth}{c YYYYYY}
		\hline
		Readers & \textbf{AUC} & \textbf{ACC} & \textbf{SEN} & \textbf{SPE} & \textbf{PPV} & \textbf{NPV} \\
		\hline
		
		R1 	
		& \makecell{0.737 \\(0.691-0.782)}	
		& \makecell{0.730 \\(0.680-0.780)}	
		& \makecell{0.564 \\(0.490-0.641)}	
		& \makecell{0.910 \\(0.860-0.953)}	
		& \makecell{0.871 \\(0.804-0.934)}	
		& \makecell{0.658 \\(0.591-0.722)}\\
		
		R1 w/ IBUS	
		& \makecell{0.793 \\(0.748-0.835)}	
		& \makecell{0.790 \\(0.743-0.833)}	
		& \makecell{0.718 \\(0.641-0.785)}	
		& \makecell{0.868 \\(0.807-0.921)}	
		& \makecell{0.855 \\(0.792-0.911)}	
		& \makecell{0.740 \\(0.671-0.802)}\\
		
		R2 	
		& \makecell{0.702 \\(0.653-0.750)}	
		& \makecell{0.697 \\(0.643-0.747)}	
		& \makecell{0.577 \\(0.500-0.654)}	
		& \makecell{0.826 \\(0.761-0.888)}	
		& \makecell{0.783 \\(0.710-0.858)}	
		& \makecell{0.643 \\(0.572-0.711)}\\
		
		R2 w/ IBUS	
		& \makecell{0.766 \\(0.720-0.811)}	
		& \makecell{0.763 \\(0.717-0.810)}	
		& \makecell{0.705 \\(0.634-0.774)}	
		& \makecell{0.826 \\(0.761-0.887)}	
		& \makecell{0.815 \\(0.752-0.879)}	
		& \makecell{0.721 \\(0.655-0.786)}\\
		
		R3 	
		& \makecell{0.755 \\(0.704-0.802)}	
		& \makecell{0.753 \\(0.703-0.800)}	
		& \makecell{0.724 \\(0.653-0.794)}	
		& \makecell{0.785 \\(0.717-0.851)}	
		& \makecell{0.785 \\(0.716-0.853)}	
		& \makecell{0.724 \\(0.658-0.794)}\\
		
		R3 w/ IBUS	
		& \makecell{0.764 \\(0.717-0.810)}	
		& \makecell{0.763 \\(0.717-0.810)}	
		& \makecell{0.737 \\(0.667-0.805)}	
		& \makecell{0.792 \\(0.725-0.858)}	
		& \makecell{0.793 \\(0.725-0.861)}	
		& \makecell{0.735 \\(0.671-0.805)}\\
		
		R4 	
		& \makecell{0.696 \\(0.644-0.746)}	
		& \makecell{0.693 \\(0.640-0.743)}	
		& \makecell{0.628 \\(0.553-0.704)}	
		& \makecell{0.764 \\(0.691-0.831)}	
		& \makecell{0.742 \\(0.667-0.814)}	
		& \makecell{0.655 \\(0.582-0.725)}\\
		
		R4 w/ IBUS	
		& \makecell{0.702 \\(0.649-0.753)}	
		& \makecell{0.700 \\(0.647-0.750)}	
		& \makecell{0.647 \\(0.571-0.724)}	
		& \makecell{0.757 \\(0.682-0.826)}	
		& \makecell{0.743 \\(0.669-0.814)}	
		& \makecell{0.665 \\(0.593-0.733)}\\
		
		R5 	
		& \makecell{0.776 \\(0.729-0.824)}	
		& \makecell{0.777 \\(0.730-0.823)}	
		& \makecell{0.782 \\(0.715-0.846)}	
		& \makecell{0.771 \\(0.700-0.839)}	
		& \makecell{0.787 \\(0.723-0.851)}	
		& \makecell{0.766 \\(0.694-0.833)}\\
		
		R5 w/ IBUS
		& \makecell{0.800 \\(0.755-0.846)}	
		& \makecell{0.800 \\(0.757-0.847)}	
		& \makecell{0.801 \\(0.735-0.862)}	
		& \makecell{0.799 \\(0.731-0.863)}	
		& \makecell{0.812 \\(0.752-0.871)}	
		& \makecell{0.788 \\(0.721-0.851)}\\
		
		R6 	
		& \makecell{0.763 \\(0.713-0.810)}	
		& \makecell{0.763 \\(0.713-0.810)}	
		& \makecell{0.763 \\(0.693-0.827)}	
		& \makecell{0.764 \\(0.690-0.831)}	
		& \makecell{0.778 \\(0.710-0.842)}	
		& \makecell{0.748 \\(0.674-0.817)}\\
		
		R6 w/ IBUS	
		& \makecell{0.823 \\(0.780-0.863)}	
		& \makecell{0.823 \\(0.780-0.863)}	
		& \makecell{0.827 \\(0.767-0.882)}	
		& \makecell{0.819 \\(0.750-0.881)}	
		& \makecell{0.832 \\(0.773-0.888)}	
		& \makecell{0.814 \\(0.752-0.873)}\\
		
		R7 	
		& \makecell{0.776 \\(0.731-0.819)}	
		& \makecell{0.770 \\(0.723-0.817)}	
		& \makecell{0.635 \\(0.561-0.707)}	
		& \makecell{0.917 \\(0.868-0.959)}	
		& \makecell{0.892 \\(0.832-0.945)}	
		& \makecell{0.698 \\(0.630-0.761)}\\
		
		R7 w/ IBUS	
		& \makecell{0.795 \\(0.752-0.836)}	
		& \makecell{0.790 \\(0.743-0.837)}	
		& \makecell{0.667 \\(0.591-0.738)}	
		& \makecell{0.924 \\(0.877-0.966)}	
		& \makecell{0.904 \\(0.847-0.955)}	
		& \makecell{0.719 \\(0.652-0.782)}\\
		
		R8 	
		& \makecell{0.833 \\(0.790-0.874)}	
		& \makecell{0.830 \\(0.787-0.873)}	
		& \makecell{0.763 \\(0.696-0.825)}	
		& \makecell{0.903 \\(0.850-0.950)}	
		& \makecell{0.895 \\(0.840-0.946)}	
		& \makecell{0.778 \\(0.715-0.839)}\\
		
		R8 w/ IBUS	
		& \makecell{0.858 \\(0.819-0.896)}	
		& \makecell{0.857 \\(0.817-0.893)}	
		& \makecell{0.814 \\(0.755-0.872)}	
		& \makecell{0.903 \\(0.846-0.947)}	
		& \makecell{0.901 \\(0.848-0.946)}	
		& \makecell{0.818 \\(0.757-0.876)}\\\hline
	\end{tabularx}
	\vspace{-2em}
	\parbox{\linewidth}{
		\raggedright
		\footnotesize
		\setlength{\baselineskip}{8pt}
		Numbers within parentheses indicate 95\% Confidence Intervals.
	}
\end{table}

\newpage

\begin{figure}[t]
	\centering
	\includegraphics[width=1\textwidth]{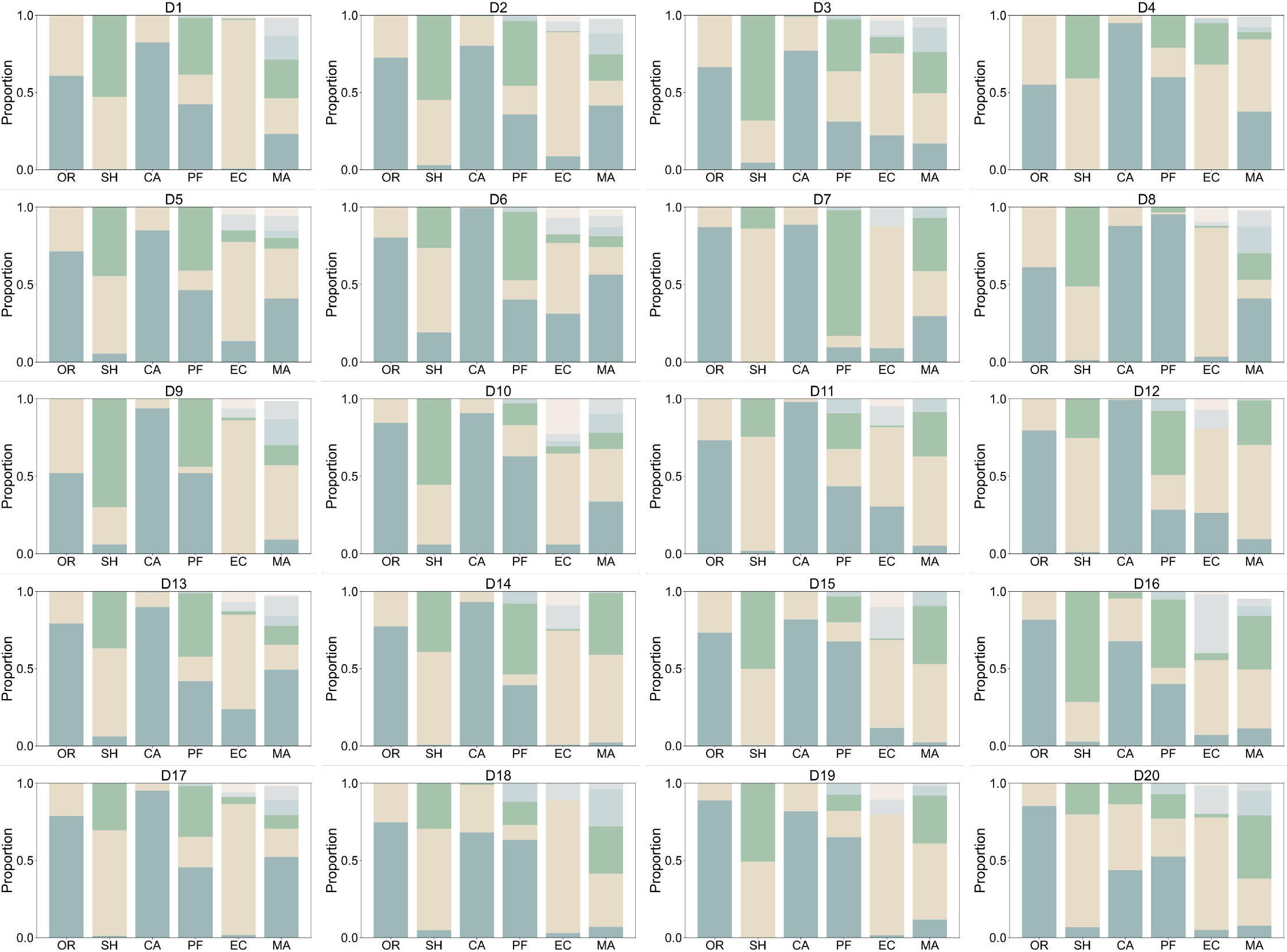}
	\caption{The distribution of BI-RADS descriptors from datasets D1-D20.
	}\label{app:text_distribution1}
\end{figure}

\begin{figure}[t]
	\centering
	\includegraphics[width=1\textwidth]{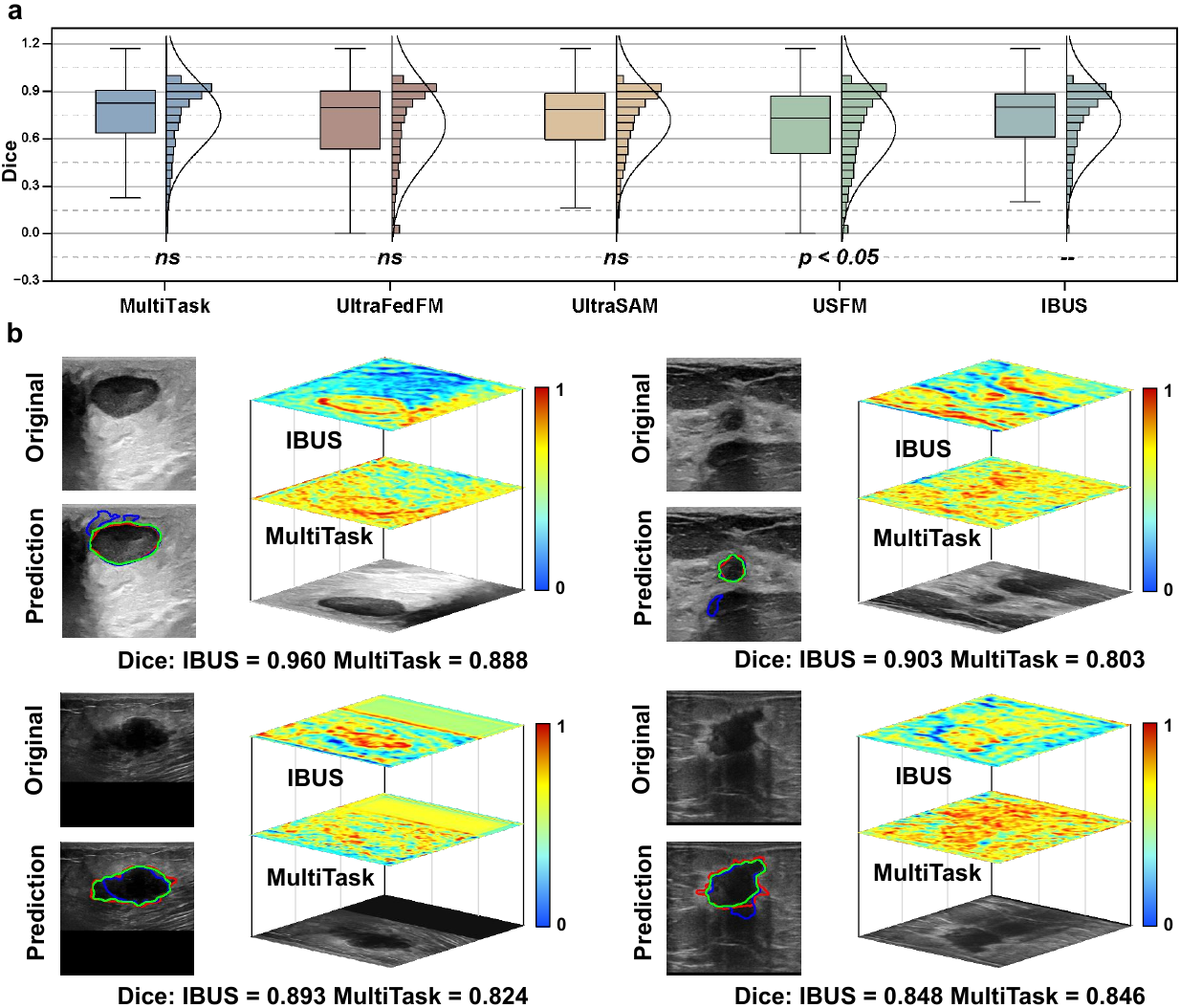}
	\caption{\textbf{Performance and visualization on lesion segmentation}.
		\textbf{a}, Segmentation performance of IBUS, MultiTask, UltraFedFM, UltraSAM, and USFM across the test sets. 
		\textbf{b}, Segmentation results and convolutional feature responses from IBUS and MultiTask of four representative cases. Red, green, and blue contours denote the ground-truth, IBUS, and MultiTask boundaries, respectively. 
		Despite using semi-supervised setting, IBUS achieved overall comparable performance with MultiTask.
		Meanwhile, IBUS showed better performance in delineating indistinct lesion boundaries, in which edge-enhanced convolution generated more concentrated boundary responses than standard convolution in MultiTask.
	}\label{app:seg_performance}
\end{figure}

\begin{figure}[t]
	\centering
	\includegraphics[width=1\textwidth]{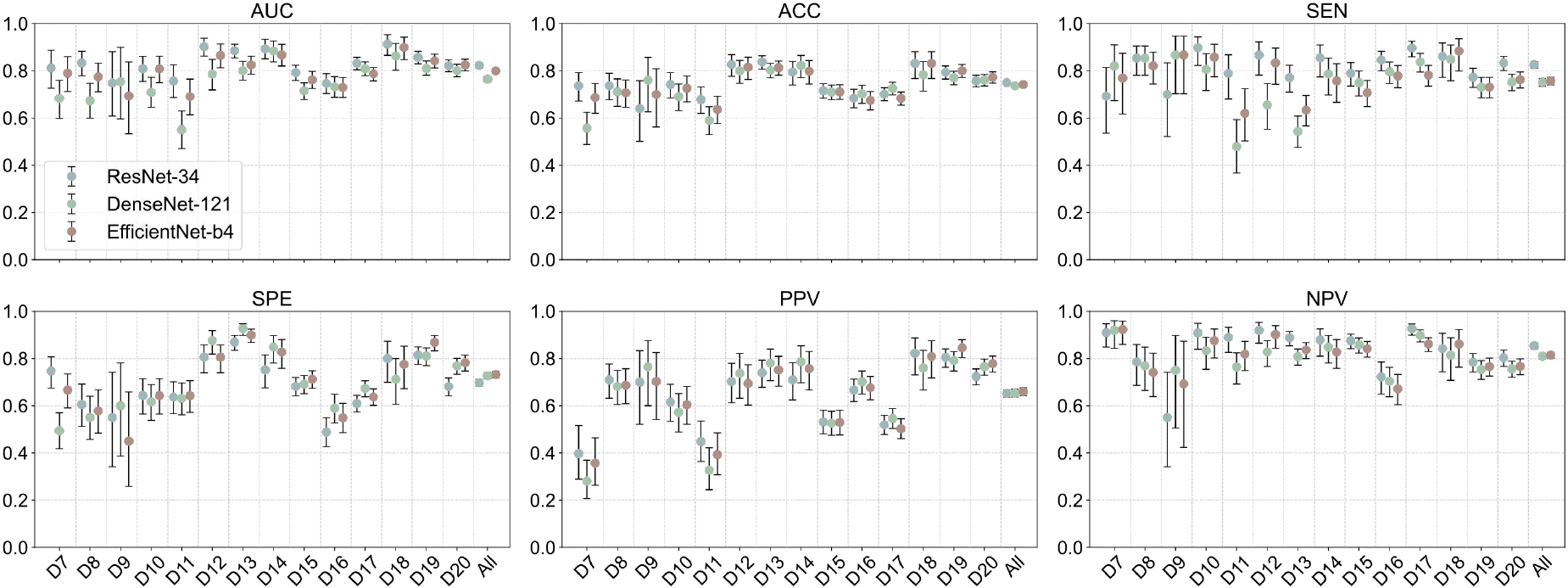}
	\caption{\textbf{Ablation study on visual extractor.}
		IBUS adopted three different feature extractors to build the models, including: 1) ResNet-34; 2) DenseNet-121; 3) EfficientNet-b4. All three backbones achieved comparable performance, indicating that IBUS is insensitive to feature extractor. Among them, ResNet-34 yielded slightly superior performance and was therefore adopted in all subsequent experiments.
	}\label{app:abla_study}
\end{figure}

\begin{figure}[t]
	\centering
	\includegraphics[width=0.98\textwidth]{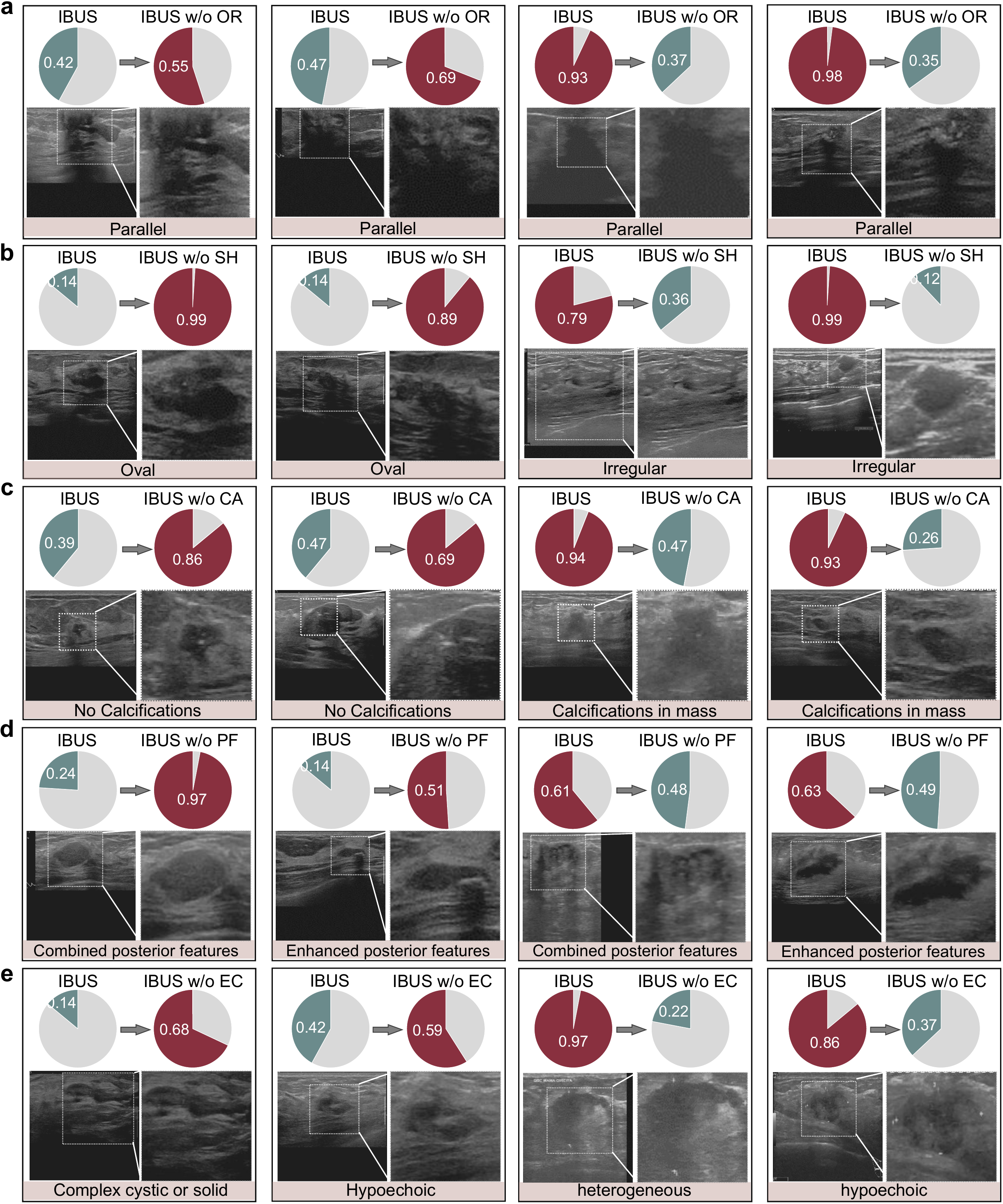}
	\caption{\textbf{Visualization on IBUS w/o OR, IBUS w/o SH, IBUS w/o PF and IBUS w/o EC.}
		\textbf{a}, Four representative misclassified cases from IBUS w/o OR.
		\textbf{d}, Four representative misclassified cases from IBUS w/o SH.
		\textbf{c}, Four representative misclassified cases from IBUS w/o CA.
		\textbf{d}, Four representative misclassified cases from IBUS w/o PF.
		\textbf{e}, Four representative misclassified cases from IBUS w/o EC.
		For IBUS w/o OR, lesions labeled as parallel can be easily misjudged because their boundaries are unclear due to echo attenuation. 
		In IBUS w/o SH, benign lesions with oval appearance are hard to distinguish because of fuzzy borders or uneven echoes, while small irregular malignant cases are easily overlooked.
		Removing calcifications descriptors substantially increased confusion between benign lesions with tiny hyperechoic areas and malignant lesions with microcalcifications.
		Obviously, the removal of posterior features makes it difficult to distinguish ambiguous enhancement/attenuation.
		As see in the cases from IBUS w/o EC, hypoechoic lesions might show similar echo with heterogenous or complex cystic or solid lesions. }\label{app:bi-rads_attribute_influence}
\end{figure}

\begin{figure}[t]
	\centering
	\includegraphics[width=1\textwidth]{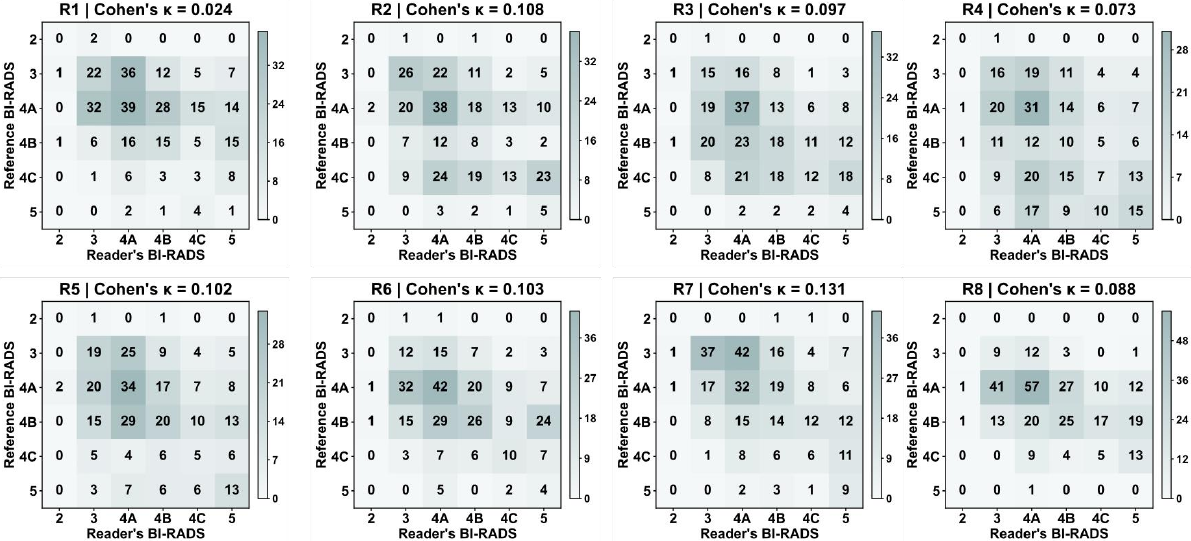}
	\caption{Risk stratification confusion matrices between reference BI-RADS grades and readers' assigned grades. Across all readers, BI-RADS assessments were predominantly concentrated between BI-RADS 3 and BI-RADS 4C. 
		Misclassifications mainly occurred between adjacent categories, for example, BI-RADS 4A lesions being assigned to BI-RADS 3 or BI-RADS 4B. Overall, grading concordance improve consistently with increasing reader's experience.
	}\label{app:reader_study_IBUS_birads_grading}
\end{figure}

\begin{figure}[t]
	\centering
	\includegraphics[width=1\textwidth]{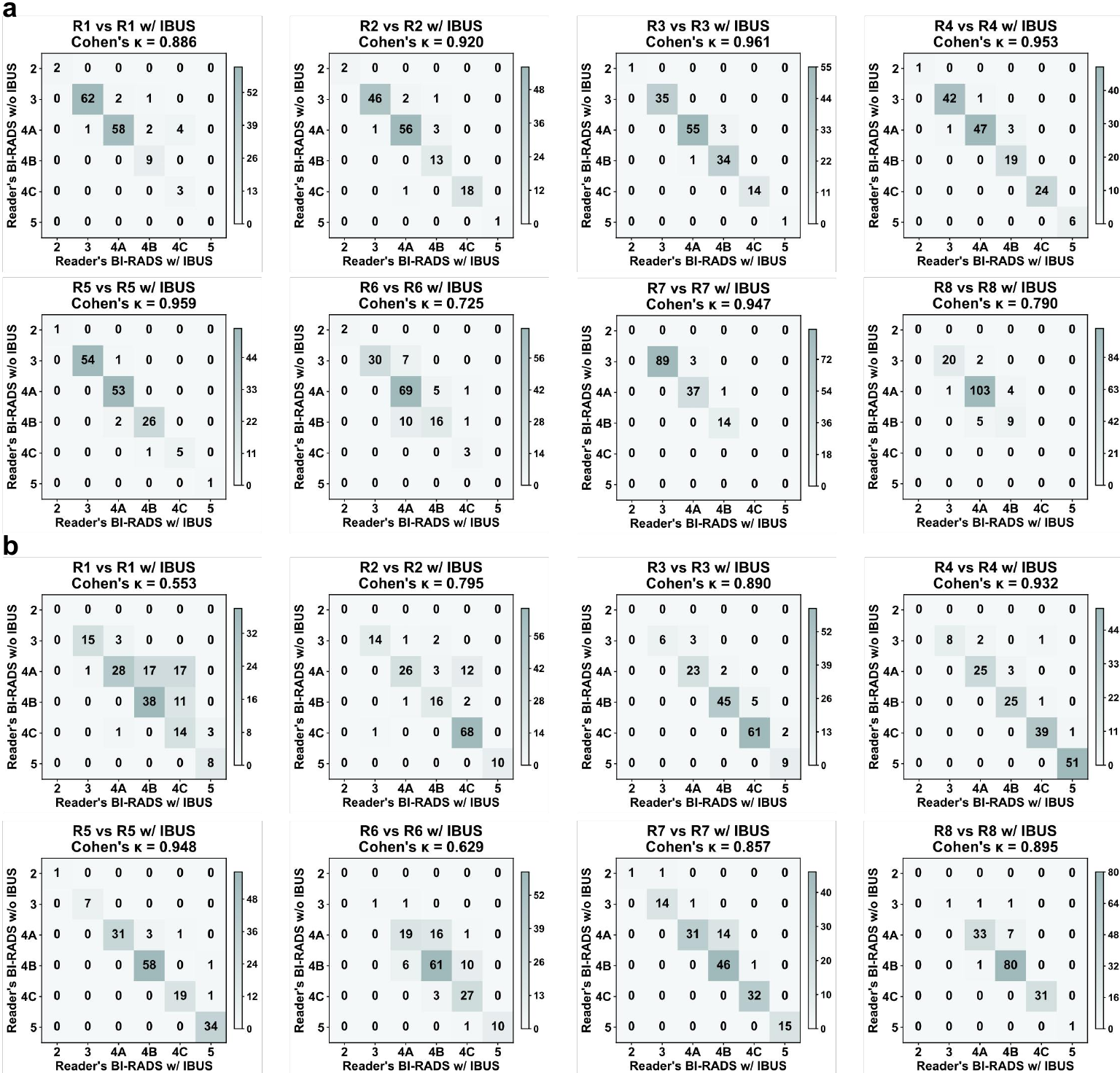}
	\caption{Risk stratification confusion matrices for benign (\textbf{a}) and malignant cases (\textbf{b}) in reader study. Comparisons were performed between readers' assessments without and with IBUS assistance (denoted as ``R* \textit{vs}. R* w/ IBUS'').
		Both IBUS and readers mainly categorized benign lesions into BI-RADS 3-4A, while malignant lesions into BI-RADS 4B-5. 
		Meanwhile, the mean kappa between readers' independent and IBUS-assisted assessments was 0.893 (range, 0.725--0.961) and 0.812 (range, 0.553--0.948), respectively.
		Importantly, no obvious grading errors was introduced with IBUS assistance, such as upgrading BI-RADS 2 lesions to BI-RADS 5 or downgrading BI-RADS 5 lesions to BI-RADS 2.
	}\label{app:reader_study_IBUS_birads_grading_mal_ben}
\end{figure}

\begin{figure}[t]
	\centering
	\includegraphics[width=1\textwidth]{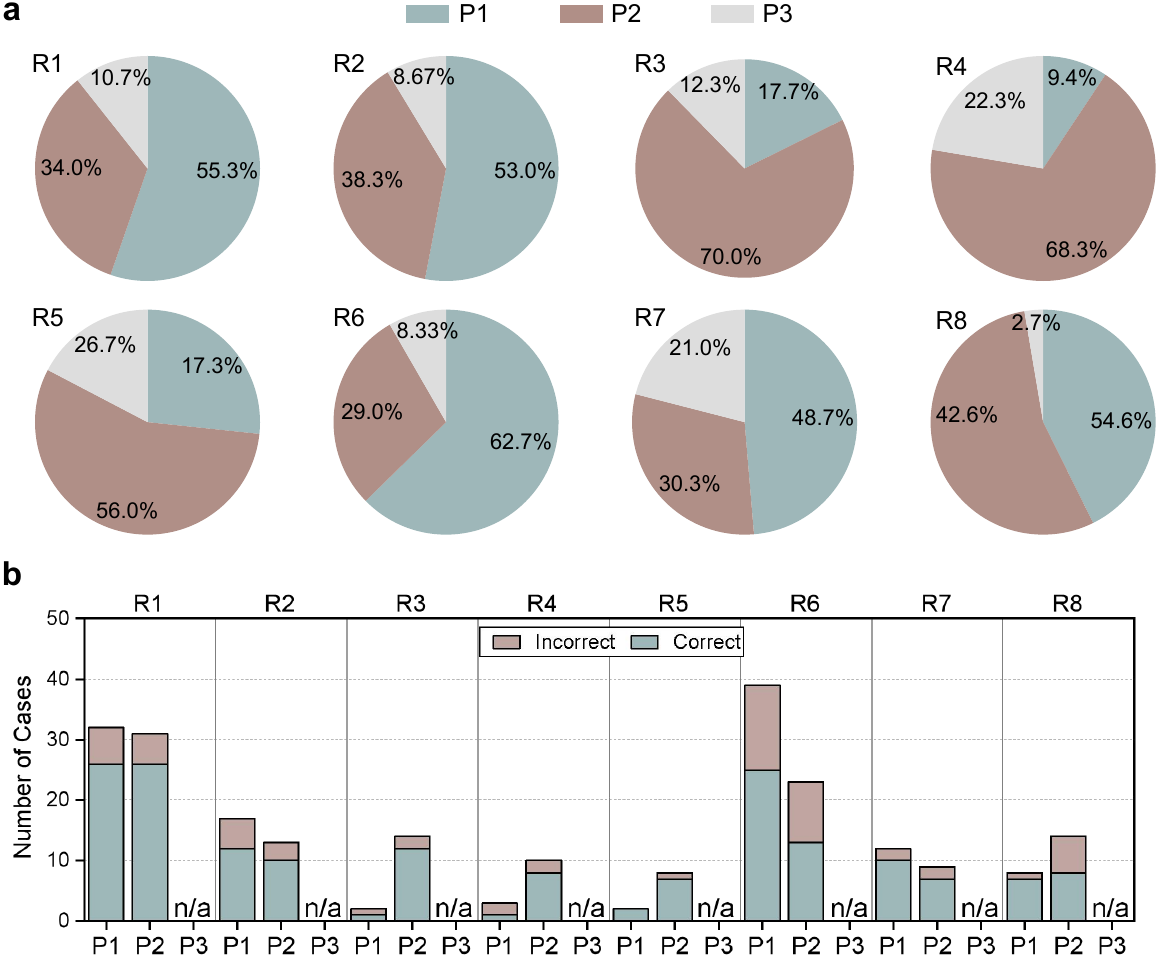}
	\caption{Evaluation results of CAMs and distributions of modified cases in P1\textasciitilde P3. 
		\textbf{a}, Evaluation results of CAMs from eight readers.
		Across all eight readers, more than 70\% of cases were assigned to P1 or P2, indicating that the model-generated CAMs showed substantial overlap with clinically relevant lesion regions in the majority of cases. These results suggest that IBUS generally localized its attention to lesion-associated regions, providing visually interpretable evidence for its predictions.
		\textbf{b}, Distribution of cases with modified BI-RADS grade in P1\textasciitilde P3.
		Notably, cases in which readers modified their BI-RADS assessments after IBUS assistance were concentrated in P1 and P2, whereas few modifications occurred in P3. 
		This pattern suggests that changes in risk stratification were more likely to occur when the model's attention maps were considered clinically plausible by the readers. 
		Rather than indicating uniform acceptance of model predictions, these findings are consistent with readers selectively incorporating IBUS outputs when the accompanying visual evidence was considered informative, potentially providing a safeguard against inappropriate reliance on model predictions.
	}\label{app:reader_study_IBUS_cam}
\end{figure}

\begin{figure}[t]
	\centering
	\includegraphics[width=1\textwidth]{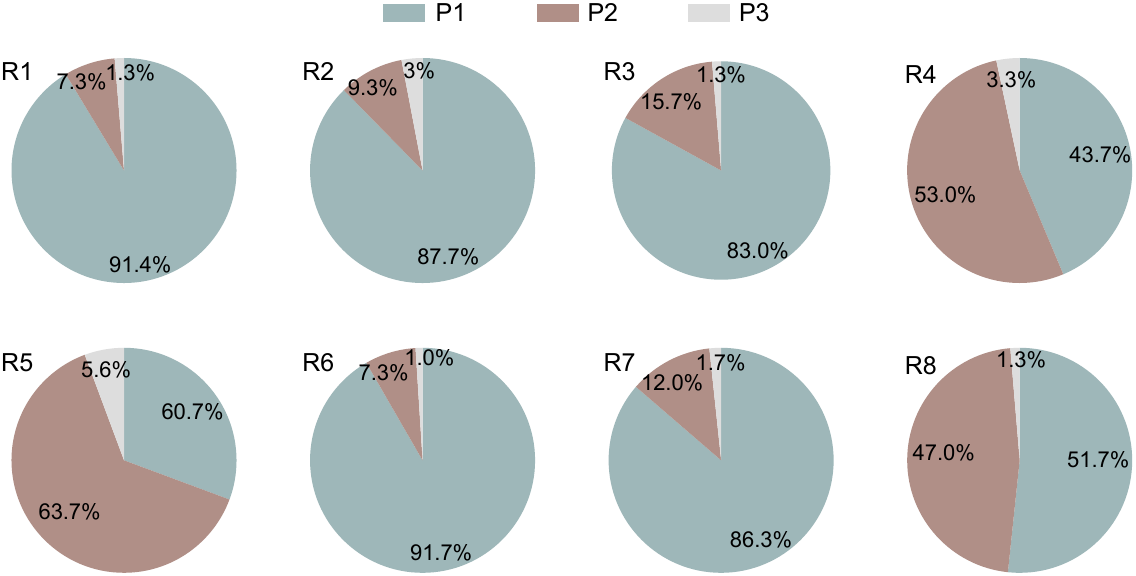}
	\caption{Evaluation of lesion localization maps by eight readers. 
		Across all readers, more than 95\% of cases were assigned to P1 or P2, indicating that the predicted localization maps covered more than 50\% of the lesion region in the vast majority of cases. 
		These findings suggest that IBUS can effectively localize breast lesions and provide spatially relevant evidence for subsequent diagnostic assessment.
	}\label{app:Lesion_localization_evaluation}
\end{figure}

\begin{figure}[t]
	\centering
	\includegraphics[width=1\textwidth]{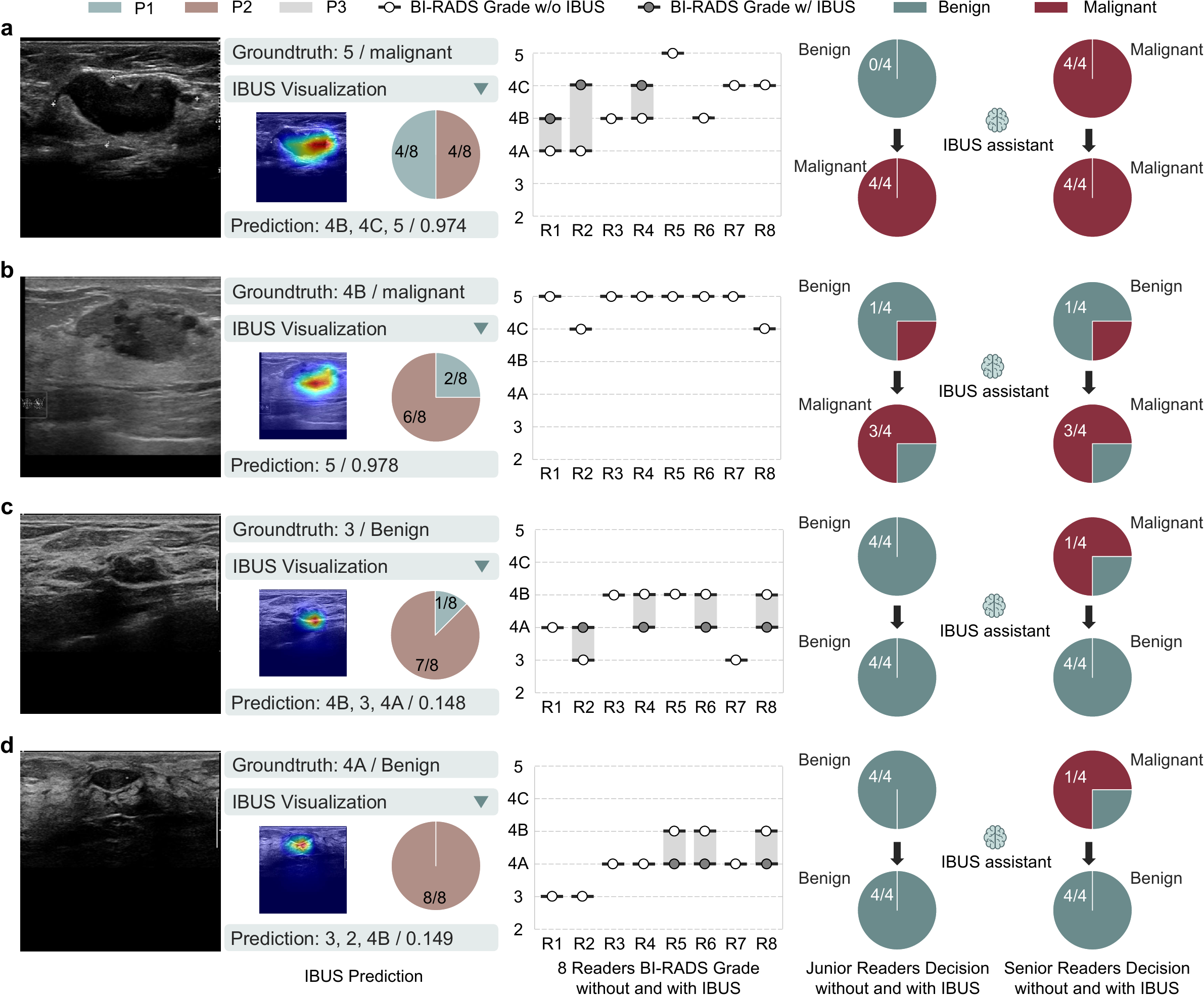}
	\caption{\textbf{Examples of IBUS prediction basis in reader study.} Heatmap analysis of four controversial cases. The red areas indicate a high level of model attention, while the blue areas indicate a low level of model attention. 
		\textbf{a}, IBUS correctly identified the lesion as malignant. In the second assessment conducted 2 weeks later with IBUS assistance, all readers think the model's attention focus on wrong region, and four out of four junior readers who initially diagnosed it as benign changed their assessment to malignant. Meanwhile, three out four junior readers raise the BI-RADS grades.
		\textbf{b}, IBUS correctly identified the lesion as malignant. In the second assessment with IBUS assistance, all readers think the model's attention focus on wrong region, and 4 out of eight readers (2 out of four junior readers and 2 out of four senior readers) who initially diagnosed it as benign changed their assessment to malignant. All readers maintain the initial BI-RADS grades.
		\textbf{c}, AI correctly identified the lesion as benign. In the second assessment with IBUS assistance, all readers think the model's attention focus on wrong region, and three out of four senior readers who initially diagnosed it as malignant changed their assessment to benign. Meanwhile, four out of eight readers (two junior and two senior readers) lower the BI-RADS grades.
		\textbf{d}, AI correctly identified the lesion as benign. In the second assessment with IBUS assistance, all readers think the model's attention focus on wrong region, and three senior readers who initially diagnosed it as malignant changed their assessment to benign. Meanwhile, three out of four senior readers lower the BI-RADS grades.
	}\label{app:reader_study_vis}
\end{figure}

\begin{figure}[t]
	\centering
	\includegraphics[width=1\textwidth]{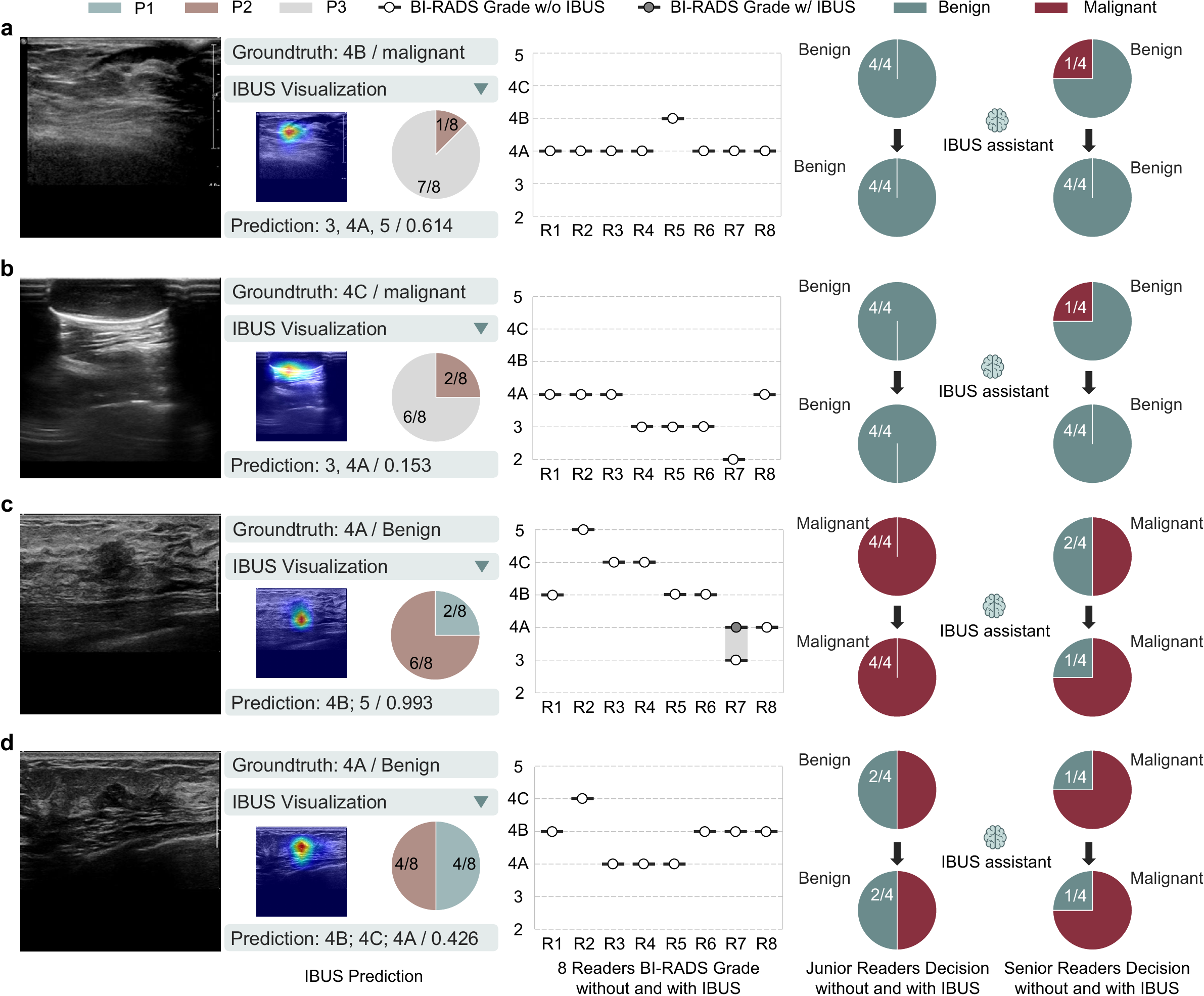}
	\caption{\textbf{Some representative wrong examples in reader study.} Heatmap analysis of four controversial cases. 
		The red areas indicate a high level of model attention, while the blue areas indicate a low level of model attention. 
		\textbf{a}, IBUS correctly identified the lesion as malignant. 
		In the second assessment conducted 2 weeks later with IBUS assistance, seven out of eight readers think the model's attention focus on wrong region, and seven out of eight readers (4 out of four junior readers and 3 out of four senior readers) who initially diagnosed it as benign did not follow the model's prediction to change their assessment to malignant, while one senior reader change his assessment to malignant. All readers maintain the initial BI-RADS grades.
		\textbf{b}, IBUS wrongly identified the lesion as malignant. 
		In the second assessment with IBUS assistance, six out of eight readers think the model's attention focus on wrong region, and 6 out of eight readers (4 out of four junior readers and 2 out of four senior readers) who initially diagnosed it as benign did not follow the model's prediction to change their assessment to malignant, while two senior readers change his assessment to malignant. All readers maintain the initial BI-RADS grades.
		All readers maintain the initial BI-RADS grades.
		\textbf{c}, AI wrongly identified the lesion as benign. In the second assessment with IBUS assistance, eight out of eight readers think the model's attention focus on clinically relative region, and 6 out of eight readers (4 out of four junior readers and 2 out of four senior readers) maintained their assessment as malignant, while one senior readers change his assessment to malignant. Meanwhile, one senior readers raised the BI-RADS grades from 3 to 4A.
		\textbf{d}, AI correctly identified the lesion as benign. In the second assessment with IBUS assistance, all readers think the model's attention focus on clinically relative region, and 5 out of eight readers (2 out of four junior readers and 3 out of four senior readers) maintained their assessment as malignant. All readers maintain the initial BI-RADS grades.
	}\label{app:reader_study_vis_wrong}
\end{figure}

\begin{figure}[t]
	\centering
	\includegraphics[width=1\textwidth]{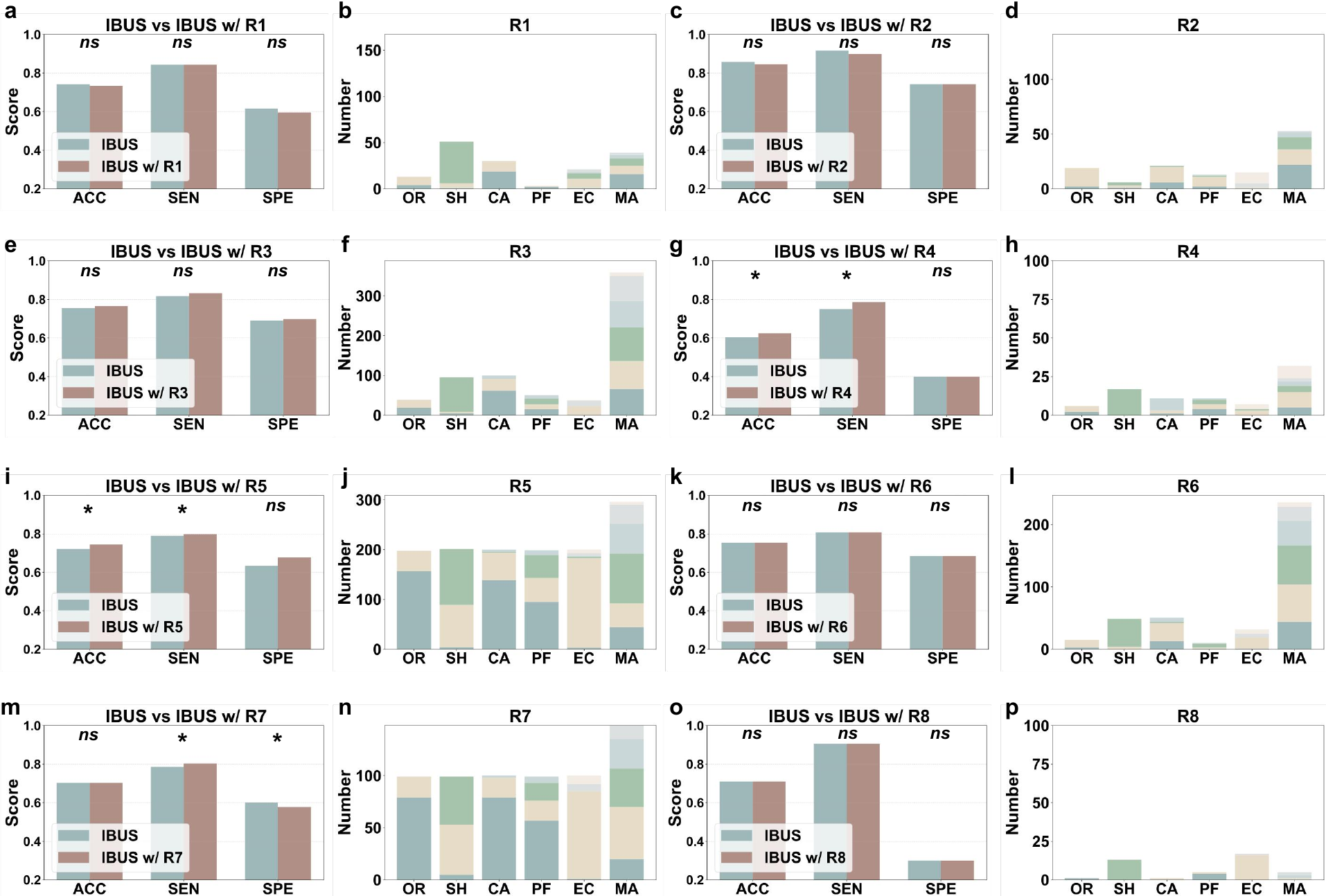}
	\caption{\textbf{BI-RADS descriptors re-evaluation in reader study}.
		Eight radiologists were invited to review the BI-RADS descriptors used by IBUS and re-annotate those they considered inaccurate or inappropriate. 
		The revised descriptors from each reader were subsequently used as inputs to IBUS (denoted as ``IBUS w/ R*'') to reassess diagnostic performance and were compared with the original IBUS results.  
		Annotation variability was observed between reader-assigned and original BI-RADS descriptors, with disagreements for most readers concentrated in the partial descriptor of cases. 
		Thus, despite descriptor revision resulted in both increases and decreases in performance across readers, most differences were not statistically significant, suggesting a degree of tolerance to inter-reader variability in BI-RADS descriptor assessment.
	}\label{app:reader_study_IBUS_revise_text}
\end{figure}

\end{document}